\documentclass[a4paper]{cas-sc}
\usepackage[numbers]{natbib}
\usepackage{amsmath,amsfonts,amssymb}
\usepackage{graphicx}
\usepackage{color,soul,colortbl}
\usepackage{mathtools}
\usepackage{bm}
\usepackage{amsmath}
\usepackage{todonotes}
\usepackage[normalem]{ulem}
\usepackage{lineno}
\newcommand{\methodName}[0]{WP-NODE}

\newcommand{\comments}[1]{}
\newcommand{\ms}[2]{#1{\scriptsize$\;\pm\;$}#2}
\def\tsc#1{\csdef{#1}{\textsc{\lowercase{#1}}\xspace}}
\tsc{WGM}
\tsc{QE}

\begin{document}


\shorttitle{}    

\shortauthors{X. Li et~al.}  

\title{\LARGE A Weak Penalty Neural ODE for Learning Chaotic Dynamics from Noisy Time Series}



\author[1,2]{Xuyang Li}[orcid=0000-0002-6846-0906]
\cormark[1]
\ead{xuyang.li@charlotte.edu}

\author[3]{John Harlim}
\cormark[2]
\ead{jharlim@psu.edu}

\author[2]{Dibyajyoti Chakraborty}[orcid=0000-0001-9844-1020]

\author[2,4]{Romit Maulik}[orcid=0000-0001-9731-8936]
\cormark[3]
\ead{rmaulik@purdue.edu}

\affiliation[1]{organization={School of Construction, University of North Carolina at Charlotte},
            addressline={Charlotte}, 
            postcode={28223}, 
            state={NC},
            country={USA}}

\affiliation[2]{organization={College of Information Sciences and Technology, The Pennsylvania State University},
            addressline={University Park}, 
            postcode={16802}, 
            state={PA},
            country={USA}}
            
\affiliation[3]{organization={Department of Mathematics, Institute for Computational and Data Sciences, The Pennsylvania State University},
            addressline={University Park}, 
            postcode={16802}, 
            state={PA},
            country={USA}}

\affiliation[4]{organization={School of Mechanical Engineering, Purdue University},
            addressline={West Lafayette}, 
            postcode={47907}, 
            state={IN},
            country={USA}}

\begin{abstract}
The accurate forecasting of complex, high-dimensional dynamical systems from observational data is a fundamental task across numerous scientific and engineering disciplines. A significant challenge arises from noisy observations of deterministic dynamics, which severely degrade the performance of data-driven models. In chaotic dynamical systems, where small initial errors amplify exponentially, it is particularly difficult to develop a model from noisy data that achieves short-term accuracy while preserving long-term invariant properties. To overcome this, we consider the weak formulation as a complementary approach to the classical $L2$-loss function for training models of dynamical systems. We empirically verify that the weak formulation, with a proper choice of test function and integration domain, effectively filters noisy data. This insight explains why a weak form loss function is analogous to fitting a model to filtered data and provides a practical way to parameterize the weak form. Subsequently, we demonstrate how this approach overcomes the instability and inaccuracy of standard Neural ODE (NODE) in modeling chaotic systems. Through numerical examples, we show that our proposed training strategy, the Weak Penalty NODE, is computationally efficient, solver-agnostic, and yields accurate and robust forecasts across benchmark chaotic systems and a real-world climate dataset.
\end{abstract}




\maketitle

\section{Introduction}
The accurate prediction of complex dynamical systems is of vital importance to science and engineering~\cite{ghadami2022data}, with applications ranging from weather forecasting and climate modeling to fluid dynamics and systems biology.
In contrast, high-fidelity simulations are computationally expensive and often diverge from real-world data. 
Data-driven methods~\cite{brunton2022data, floryan2022data, kaptanoglu2021promoting}, particularly those based on machine learning~\cite{legaard2023constructing, gilpin2021chaos, HJLY:19}, have emerged as a powerful tool for learning these system dynamics directly from observational data and offer a computationally efficient alternative to first-principles simulation.

Numerous machine learning models have demonstrated strong performance in forecasting complex spatiotemporal dynamics. These include recurrent models such as Long Short-Term Memory (LSTM) \cite{HJLY:19,chattopadhyay2020data}, reservoir computing (RC) \cite{kobayashi2021dynamical, yan2024emerging, bianchi2020reservoir}, temporal convolutional networks (TCNs) \cite{lea2016temporal, perumal2023temporal}, attention-based Transformers~\cite{vaswani2017attention, he2025chaos, patil2023autoregressive}, and more recently, random feature maps (RFMs)~\cite{mandal2025learning}. Another prominent approach is the Neural Ordinary Differential Equation (Neural ODE) \cite{chen2018neural, oh2025comprehensive}, which models dynamics by parameterizing the vector field of an ordinary differential equation with a neural network. Although these models are effective when trained on clean data, their forecasting capabilities deteriorate significantly when the training data is corrupted by noise~\cite{goyal2023neural,gottwald2021supervised}. This is particularly problematic in chaotic systems, where small initial errors can amplify exponentially and lead to forecast failure \cite{mandal2025learning}.

Various methods have been proposed to learn from noisy data{\color{black}, where the dynamics are
deterministic and the noise enters only through the observations}. For instance, Bayesian and probabilistic models, such as Gaussian Process-based approaches \cite{yang2021inference}, Bayesian neural surrogates \cite{seleznev2019bayesian, yang2021b}, {\color{black}and Bayesian system identification~\cite{galioto2024bayesian}}, offer principled uncertainty quantification but often struggle with high computation costs, sensitivity to hyperparameters, restrictive modeling assumptions, and costly inference.
Beyond machine learning, Kalman filtering and various data assimilation techniques have been proposed to denoise initial conditions and estimate parameters of dynamical systems \cite{cheng2023machine}. More recently, Kalman filtering has also been used to learn dynamical systems \cite{gottwald2021supervised}. While this approach is competitive, as we will show, it does not scale well to high-dimensional problems unless additional empirical parameterizations are embedded within the framework.

Classic Galerkin-based methods introduce weak formulations by projecting residuals onto a test function space, which naturally suppresses high-frequency noise when an appropriate test function space is chosen. This weak form concept has inspired foundational work in learning nonlinear PDEs \cite{gurevich2019robust} and broader models \cite{stephany2024weak, bortz2024weak}. These include weak form or variational physics-informed neural networks (PINNs) \cite{kharazmi2021hp} and weak-SINDy \cite{messenger2021weak}, which improve robustness by integrating residuals over subdomains.

Building on these encouraging results, the present paper demonstrates that the weak formulation effectively filters noise. Furthermore, while the SINDy modeling approach assumes that the underlying vector space of the ODEs lies within its model class (i.e., the linear span of its dictionary), we aim to demonstrate the effectiveness of weak form without this assumption. Specifically, we seek to understand how training NODEs with weak form loss functions can help {\color{black}robustly recover complex ODE vector fields from noisy data, without the restrictive assumption that the dynamics lie within the linear span of a predefined basis dictionary}. A relevant recent work is the variational formulation-based NODE (VF-NODE)~\cite{zhao2025accelerating}, {\color{black}which also trains NODEs with a weak (variational) loss, using a global Fourier sine basis together with a spline representation of the noisy trajectory. Because the resolution of this global basis is tied to the length of the training time series, its cost grows accordingly on the long time series required for learning chaotic dynamics; a quantitative comparison is provided in Appendix~\ref{app:vfnode}. In contrast, \methodName{} avoids prior data interpolation by integrating directly over raw noisy observations, employs locally supported polynomial test functions rather than a global Fourier basis, and combines the weak form with a short-horizon strong component to stabilize optimization in chaotic regimes.}

Our motivation for using NODEs in this study, beyond their fundamental design for approximating ODE vector fields, is as follows. The continuous formulation of NODEs enables predictions at arbitrary times. Furthermore, NODEs provide significant architectural flexibility; their dynamics can be modeled by a range of structures, from simple feedforward networks to more complex architectures like convolution, U-Net \cite{kidger2022neural, ronneberger2015u}, or transformers. This allows them to capture intricate spatial and temporal dependencies that are beyond the scope of traditional recurrent models.

On the flip side, our choice to focus on NODEs is also driven by the practical challenges of training an effective model with high predictive skill for chaotic dynamical systems, even with noise-free training data. This difficulty is largely due to issues with the standard training procedure. The adjoint sensitivity method~\cite{chen2018neural}, commonly used for NODE training, is memory-efficient but incurs a high computational cost due to repeated forward and backward ODE solves. Moreover, the backward integration step is often unstable, leading to exploding or vanishing gradients \cite{chakraborty2024divide, linot2023stabilized}, particularly in stiff systems \cite{fronk2025training}. This computational overhead makes it prohibitive to train a NODE model whose long-horizon rollouts, i.e., recursive predictions over extended durations, fit to a long time series of training data, which is critical for accurately predicting chaotic dynamics.  While additional architectural constraints or specialized training techniques, such as the multistep penalty formulation in MP-NODE \cite{chakraborty2024divide}, can yield stable predictions, they are computationally expensive and often lack predictive accuracy.

This study aims to determine if the weak form approach can overcome the practical issues that hinder NODEs from making accurate short- and long-term predictions of chaotic dynamical systems. {\color{black}Particularly, we will verify that robust and accurate NODE models can be achieved by a simple modification in the training procedure. That is, by adding the weak form as a penalty in the training loss. We call the resulting model \methodName{}.} Furthermore, we will investigate whether the  \methodName{} can produce an NODE model that is robust across various numerical solvers, addressing a key limitation of standard training procedures.

The remainder of this paper is organized as follows. In Section~\ref{sec:method}, we give a short overview of training NODEs with noisy data, define and clarify how the weak formulation is effectively filtering noisy data, introduce a parameter estimation strategy for forming the appropriate weak loss functions, and introduce the Weak Penalty Neural ODE (\methodName{}). 
Section~\ref{sec3} details the experimental setup, including NODE parameterization, training procedures, and the metrics used for evaluating prediction accuracy. In Section~\ref{sec4}, we present a comparative study using the low-dimensional Lorenz-63 model. Section~\ref{sec5} extends this analysis to higher-dimensional systems, including the Lorenz-96 and Kuramoto-Sivashinsky equations, as well as the ERA5 atmospheric dataset. Finally, we conclude with a summary of our findings. {\color{black}Five} appendices provide supplementary details on the Lorenz-63, Kuramoto–Sivashinsky, and ERA5 results, the quadrature approximation used for the weak form loss function, {\color{black}and a comparison with VF-NODE.}

\section{Learning Dynamical Systems with Neural ODEs}\label{sec:method}
For the discussion below, assume a noisy time series is given,
\begin{equation}
v_n = u_n + \xi_n, \quad \xi_n \sim \mathcal{N}(0,\sigma^2), \quad n =0,\ldots, N-1,  \label{noisyobs} 
\end{equation}
where $u_n = u(t_n)$ is a solution of a system of unknown initial value problems,

\begin{equation}
\frac{du}{dt} = u'= f^\dagger (u,t), \quad u(t_0) = u_0,    \label{truedynamics}
\end{equation}
at discrete time steps $\{t_n\}_{n=0}^{N-1}$, which is a set of $N$ evenly spaced time points within a training window, satisfying $t_n = t_0 + n \Delta t$. As formulated in Eq. \eqref{noisyobs}, the state $u_n$ is corrupted by an independent and identically distributed (i.i.d.) noise $\xi_n$ of an unbiased
Gaussian distribution, with standard deviation set as
$$ \sigma = \sigma_{\text{NR}} \text{RMS}(u),$$
where $\sigma_{\text{NR}}$ is to be specified in various numerical experiments. The goal is to learn the underlying vector field $f^\dagger$. 

Section~\ref{sec21} provides a brief overview of the NODE and numerical issues encountered in training this class of models. Section~\ref{sec22} discusses the weak formulation and explains how this formulation effectively filters the noise. Section~\ref{sec23} presents a simple strategy to accelerate the NODE training procedure by incorporating the weak form as a penalty to the NODE loss function.

\subsection{Neural Ordinary Differential Equations.}\label{sec21}
NODEs \cite{chen2018neural} generalize residual networks by replacing discrete-layer architectures with a continuous-time dynamics model. Given an initial condition \(u(t_0) = u_0\), the evolution of the state is governed by:
\begin{align}\label{eq:node}
    \hat{u}'(t;\theta) = f(\hat{u}(t;\theta), t; \theta),
\end{align}
where \(f\) is a neural network parameterized by \(\theta\). The solution at a later time \(t\) is determined by solving this initial value problem using standard adaptive ODE solvers (e.g., Runge–Kutta, Dormand–Prince)~\cite{dormand1980family, butcher2016numerical}, which adjust their step size to balance computational cost and precision. 

The standard training paradigm for a NODE is the strong formulation, 
which enforces that the predicted trajectory $\hat{u}(t;\theta)$ closely matches the ground-truth state $u_n = u(t_n)$ at discrete observation times, assuming their availability. In this paper, noisy observations are considered as in Eq. \eqref{noisyobs}. Given $\{v_n\}_{n=0}^{T-1}$,  this objective is typically achieved by minimizing a loss function, such as the mean squared error (MSE),
\begin{align}\label{eq:strong_loss}
\mathcal{L}_{\text{strong}}(\theta) 
= \frac{1}{T-1} \sum_{n=1}^{T-1} \left\| v_n - \hat{u}(t_n; \theta) \right\|^2,
\end{align}
where $T$ denotes the trajectory window size (corresponding to $T-1$ rollout steps, see Table~\ref{tab:sup-config-merged} for values of $T$ on different examples). In this strong form baseline, the model treats any given time step $t_0$ as an initial condition and performs a rollout by recursively predicting the subsequent $T-1$ states through the ODE solver.

For stability of training and prediction with the NODE model, the data is normalized using a MinMax scaler. Given a training data set $\{v_0, \ldots, v_{N-1}\}$, a set of $\frac{N-1-T}{q_s}$ time series of length $T$ are sampled at every $q_s$ (the interval between the starting points of consecutive sequences) for training with the strong form loss function in \eqref{eq:strong_loss}. When $q_s=1$, the resulting $N-T-1$ time series consists of $\{v_0,\ldots, v_{T-1}\}, \{v_1, \ldots, v_{T}\}, \ldots, \{v_{N-T},\ldots,v_{N-1}\}$.

To optimize the parameters $\theta$ with respect to this loss, gradients are computed using direct backpropagation through the solver instead of the adjoint sensitivity method. 
Since the strong form baseline is restricted to very short horizons, direct backpropagation ensures numerical stability and computational efficiency while incurring only a negligible memory cost. However, the gradient computation becomes prohibitively expensive for chaotic or stiff dynamics, especially for large $T$, where the adjoint sensitivity method is typically required to maintain constant memory overhead. Exploding gradients and numerical instabilities demand small solver step sizes and tight tolerances, leading to high computational overhead~\cite{chakraborty2024divide, fronk2025training}. 

Beyond these practical issues, it remains unclear how the strong formulation will perform when the NODE is fitted to noisy data. These limitations motivate the following discussion.

\begin{table}[!htbp]
\centering
\small
\setlength{\tabcolsep}{4pt}
\renewcommand{\arraystretch}{1.15}
\caption{Hyperparameters for numerical simulations.}
\label{tab:sup-config-merged}
\begin{tabular}{lcccccccc}
\toprule
& & & \multicolumn{1}{c}{Strong Form} & \multicolumn{3}{c}{Weak Form} & \multicolumn{2}{c}{\methodName{}} \\
\cmidrule(lr){4-4} \cmidrule(lr){5-7} \cmidrule(lr){8-9}
System & $\Delta t(s)$ & $N$ & $T$ & $p$ & $q$ & $\ell$ & $T$ & $\lambda$\\
\midrule
\textbf{L63} & $0.01$ & $10^4$ & $25$ & $8$ & $2$ & $50$ & $2$ & 0.5\\
\textbf{L96} & $0.01$ & $10^5$ & $25$ & $8$ & $2$ & $80$ & $2-6$ & 0.5\\
\textbf{KS}  & $0.25$ & $10^5$ & $25$ \& $50$ & $16$& $2$ & $60$ & $2-6$ & 1\\
\textbf{ERA5}  & $1$ (Euler) & $11,576$ & $6$ & $16$ & $1$ & $8$ & $2$ & 1\\
\bottomrule
\end{tabular}
\end{table}

\subsection{Weak Formulation}\label{sec22}
The weak formulation enforces differential equations into an integral form by weighting it with smooth and compactly supported test functions. 
This approach avoids the need for recursive forward integration through noisy data via an ODE solver, and instead acts as a low-pass filter that naturally improves robustness to observational noise. Unlike VF-NODE~\cite{zhao2025accelerating}, which relies on global Fourier basis functions and spline pre-processing, the locally supported polynomial test functions adopted here provide a local noise-filtering mechanism, as detailed in Section~\ref{sec221}.

Let $\phi$ be a smooth test function defined on the time interval $[a,b]$, and satisfying $\phi(a)=\phi(b)=0$. The weak formulation of Eq.~\eqref{truedynamics} requires that the residual is orthogonal to all test functions under the inner product of $L^2([a,b])$,
\begin{align}\label{eq:weak_identity}
\int_{a}^{b} \Bigl(u'(t) - f^\dagger(u(t), t)\Bigr)\phi(t)dt = 0.
\end{align}
If only time series of $u_n$ are given, it is more convenient to employ integration by parts to avoid estimating the derivative $u'(t)$. Particularly, one can take the first term in Eq. \eqref{eq:weak_identity} and reformulate it as follows:
\begin{align}\label{eq:int_by_parts}
\int_{a}^{b}u'(t)\phi(t)\,dt = \Bigl[u(t)\phi(t)\Bigr]_{a}^{b} -\int_{a}^{b}u(t) \phi'(t)\,dt  = -\int_{a}^{b}u(t) \phi'(t)\,dt ,
\end{align}
where the boundary contribution  vanishes since the test function satisfies $\phi(a)=\phi(b)=0$.

The key idea of weak formulation is to approximate $f^\dagger$ with any choice of model $f$ to satisfy the constraint in Eq. \eqref{eq:weak_identity}. Together with the identity in Eq. \eqref{eq:int_by_parts}, the resulting weak approximation is,
\begin{equation}\label{weakid2}
\int_{a}^{b}\left(u(t) \phi'(t) + f(u(t),t;\theta)\phi(t)\right)\,dt = 0.
\end{equation}
In \cite{bortz2024weak}, they consider SINDy for $f$ and assume that $f^\dagger$ is in the space spanned by the SINDy dictionary. The present work removes such a strong assumption and a generic NODE model is considered for $f$. 

To enforce Eq. \eqref{weakid2}, the test function $\phi$ and a set of domains $\{[a_k,b_k]\subset \mathbb{R}: a_k,b_k \in \{t_1,\ldots,t_N\},\, k\geq 1\}$ need to be chosen for integration. The choice of domain is parameterized by the length of each interval, defined as an even integer $\ell = \frac{b_k-a_k}{\Delta t}$, and the distance $q$ between the centers of two adjacent intervals. With this definition, a set of integration domains parameterized by $\ell$ and $q$ is given by  $$\mathcal{I}_{q,\ell}:=\left\{\ldots, \underbrace{\left[t_{k-q}-\frac{\ell}{2}\Delta t,t_{k-q}+\frac{\ell}{2}\Delta t\right]}_{\smash{=I_{k-q,\ell}}}, \underbrace{\left[t_k-\frac{\ell}{2}\Delta t,t_k+\frac{\ell}{2}\Delta t\right]}_{\smash{=I_{k,\ell}}}, \underbrace{\left[t_{k+q}-\frac{\ell}{2}\Delta t,t_{k+q}+\frac{\ell}{2}\Delta t\right]}_{\smash{=I_{k+q,\ell}}}, \ldots\right\},$$ which may or may not overlap. 
{\color{black}Note that each integration domain $I_{k,\ell}$ requires at least three consecutive time points, precluding the use of isolated snapshot pairs.}

The choice of test functions is critical for the accuracy and efficiency of weak form integration.  
Following \cite{bortz2024weak}, the even polynomial family is adopted on the reference domain $\phi_p: [-1,1] \to \mathbb{R}$,
\begin{align}
\phi_p(s)=(1-s^{2})^{p},\qquad s\in[-1,1],\; p\in\mathbb{N}.\label{poly}
\end{align}
This construction ensures vanishing at the boundaries $(\phi_p(\pm 1)=0)$, allows smoothness to be tuned by the integer \(p\), and, being polynomial, permits semi-analytical evaluation of the weak form integrals. The test function on each interval $I_{k,\ell}$ is given by
\[
\phi_{p}^{k,\ell} (t): = (\phi_p \circ \tau_{k,\ell}^{-1})(t) =  \left(1-\left(\frac{2}{\ell\Delta t}(t-t_k)\right)^2\right)^p,
\]
where $\tau_{k,\ell}: [-1,1] \to I_{k,\ell}$ is a linear map defined as $t = \tau_{k,\ell}(s) =\frac{\ell\Delta t}{2}s + t_k$ for any $s\in[-1,1]$.

{\color{black}\subsubsection{Numerical weak form loss functions.} }Using the affine map $t=\tau_{k,\ell}(s)$, the integrals on each $I_{k,\ell}$ can be transformed to,
\begin{align}\label{loc_integral}
\begin{aligned}
\int_{I_{k,\ell}} u(t)(\phi^{k,\ell}_p)'(t)dt 
&= \int_{-1}^{1}  u(\tau_k(s))\phi'_p(s)ds \approx \sum_{j=-\ell/2}^{\ell/2} v_{k+j} \omega_{j} := V_{k,\ell}^p, \\
\int_{I_{k,\ell}} f(u(t),t;\theta)\phi(t)dt
&= \frac{\ell\Delta t}{2}\int_{-1}^{1} f(u(\tau_k(s)),\tau_k(s);\theta)\phi_p(s)ds \approx \frac{\ell\Delta t}{2} \sum_{j=-\ell/2}^{\ell/2} f(v_{j+k},t_{j+k};\theta)\hat{\omega}_j:= F^p_{k,\ell}(\theta),
\end{aligned}
\end{align}
Using the relations $$ds=\tfrac{2}{\ell\Delta t}\,dt \quad \text{and} \quad \frac{d\phi^{k,\ell}_p(t)}{dt} = \frac{d\phi_p(s)}{ds}\frac{ds}{dt} = \phi_p'(s)\frac{2}{\ell\Delta t},$$ these integrals are approximated with quadrature rules with appropriate precomputed weights, $\omega_j, \hat{\omega}_j$ (see Appendix~\ref{app:integration} for the detailed calculation of these weights). It is also noted that these integrals are attained by evaluating the integrand on noisy data $v_n$ since $u_n$ are not available.

Numerically, the weak form loss function is given by
\begin{equation}
\mathcal{L}_{weak}(\theta) = \sum_{k\in \mathcal{K}} \left\|V_{k,\ell}^p + F^p_{k,\ell}(\theta)\right\|^2,\label{weakloss}    
\end{equation}
which depends on three parameters $p,q,\ell$ that can be pre-determined based on the subsequent discussion. In Eq.~\eqref{weakloss}, the errors are averaged over the following set of indices where the integrals are taken over $\mathcal{I}_{q,\ell}$:
\[
\mathcal{K} = \{k = i+jq \in \mathbb{Z}: i,j\in \mathbb{Z} \text{ and } I_{i+jq,\ell} \in \mathcal{I}_{q,\ell}\}.
\]

\subsubsection{Understanding the Weak Formulation} \label{sec221}
The following analysis clarifies the role of the weak form in filtering noise, providing a justification for the use of noisy data $v_n$ in approximating the integrals in Eq.~\eqref{loc_integral}. To this end, $f(v)$ is approximated with the following finite sum,
\begin{equation}
f(v) \approx f_{\mathcal{K}} =  \sum_{k\in \mathcal{K}} \hat{f}_k \phi_p^{k,\ell},\label{approximation}   
\end{equation}
for fixed $p$ and $\ell$. {\color{black}By exploiting the weak identity in Eq.~\eqref{weakid2}, the computable $V_{i,\ell}^p$ (defined in Eq.~\eqref{loc_integral}) approximates the negative integral of the unknown vector field, $-F_{i,\ell}^p$.} Here, $\hat{f}_k$ can be obtained by solving the following linear algebra problem 
\begin{eqnarray}
V_{i,\ell}^p \approx -\langle f(v),\phi_p^{i,\ell}\rangle \approx -\sum_{k\in \mathcal{K}} \langle\phi_p^{k,\ell},\phi_p^{i,\ell}\rangle\hat{f}_k = -\sum_{k\in \mathcal{K}}A_{ik} \hat{f}_k \quad \Longleftrightarrow \quad V \approx -A \hat{f},\label{lin_alg}
\end{eqnarray}
where the inner product is defined as $\langle f,g \rangle = \int_\mathbb{R} f(t)g(t)\,dt$. If $A$ is invertible, then it is clear that $\hat{f}_k \in \text{span}\{V_{i,\ell}^p: i \in \mathcal{K}\}$ for all $k\in \mathcal{K}$. Mathematically, this identifies an isomorphism between $\{V_{i,\ell}^p\}_{i \in \mathcal{K}}$ and $\{\hat{f}_k\}_{k \in \mathcal{K}}$. This implies that fitting to $\{V_{i,\ell}^p\}_{i \in \mathcal{K}}$ as proposed in Eq.~\eqref{weakloss} is equivalent to fitting to $\{\hat{f}_k\}_{k \in \mathcal{K}}$ which is effectively
$f_\mathcal{K}$. 

{\color{black}A more thorough analysis that connects the expansion representation of the filtered solution to the filtering mechanism is reported in a recent follow-up work \cite{kreider2026learning}. In our context, if one assumes $f$ is smooth and the signal-to-noise ratio is large, $\sigma \ll 1$, then one can approximate $f(v) = f(u)+ c\sigma$, where $c=\|f'\|_{\infty}$. Subsequently, the error analysis in Section 3.3 of \cite{kreider2026learning} can be applied to the filter $f(v)$. Specifically, one deduces that the filter error is composed of bias and variance terms. The variance error term depends linearly on the number of basis functions, $|\mathcal{K}|$. While an explicit expression of the bias term on $(p,q,\ell)$ is not attained, a semi-analytical expression of the bias term (see Section 3.3 and Appendix B in \cite{kreider2026learning}) suggests that it is consistent for small $q$ and large $\ell$.}

\begin{figure}[!htbp]
    \centering
    \includegraphics[width=\textwidth]{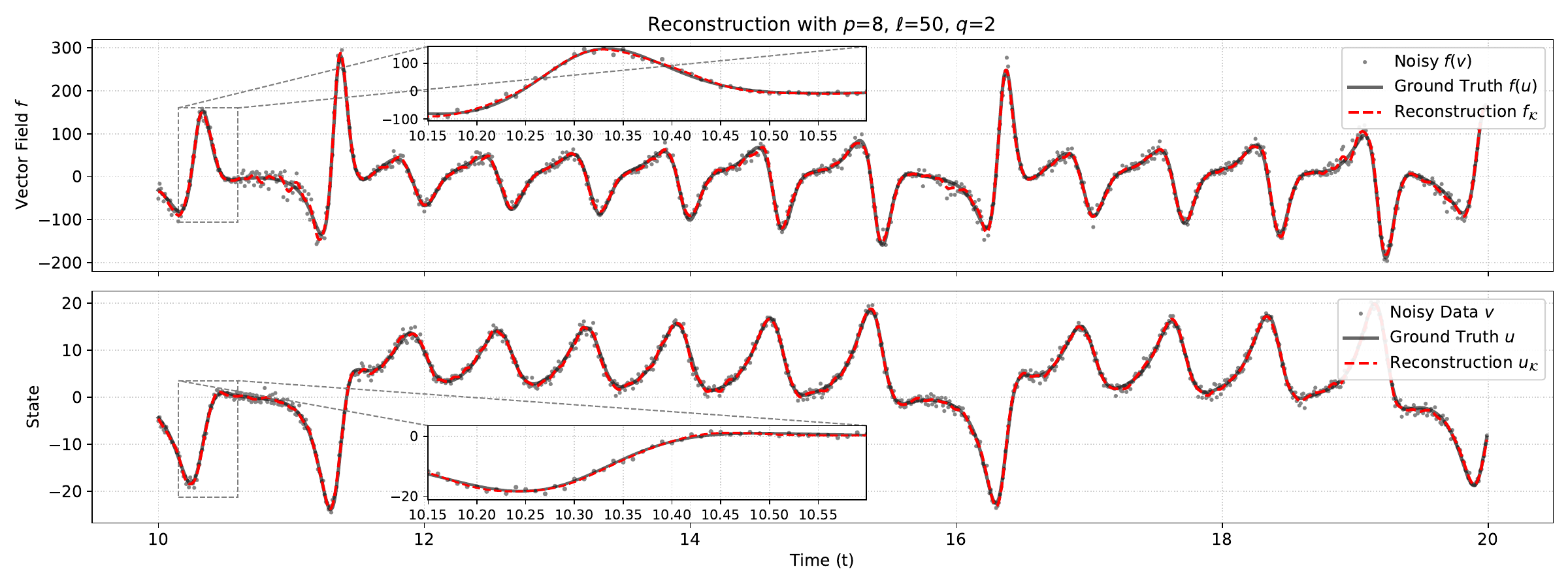}
    \caption{Reconstruction of the Lorenz-63 state and vector field, using the polynomial basis. The reconstructed signal (red dashed) effectively recovers the ground truth (black) from noisy observations (5\% noise, gray dots).
    }
    \label{fig:reconstuction}
\end{figure}

Next, numerical experiments verify that, under an appropriate choice of $(p,q, \ell)$, the finite expansion $f_\mathcal{K}$, in fact, is a filtered signal of the noisy observation data $f\circ v$. As an example, $f\circ v$ is considered to correspond to the $y$-component of the vector field of the Lorenz-63 model, $f(u) = f(x,y,z) = x(\rho-z)-y$, where the signal $u=(x,y,z)$ is corrupted by Gaussian noises with mean zero and variance $\sigma^2$. In this numerical demonstration, a $\sigma_{\text{NR}}=$5\% noise is considered. Figure~\ref{fig:reconstuction} shows the reconstructed signal $f_\mathcal{K}$ (red dashed) that looks closer to $f(u)$ (black), compared to the noisy observed data $f(v)$ (gray dots), for  $p=8, q=2, \ell = 50$. This numerical verification provides an intuitive explanation responding to the first question at the beginning of this section. Specifically, fitting to $\{V^p_{i,\ell}\}_{i\in \mathcal{K}}$ is equivalent to fitting a representation of the filtered signal, $f_\mathcal{K}$.
  
In practice, however, typically only $v$ is observed instead of $f(v)$. By repeating the same analysis in Eq.~\eqref{approximation}-\eqref{lin_alg} to remove noise in the $y$-component of the Lorenz-63 solution, that is, considering the filtered solution defined as,
\begin{eqnarray}
u_\mathcal{K} = \sum_{k\in \mathcal{K}} \hat{v}_k \phi^{k,\ell}_p, \quad\text{with coefficients that satisfy }\quad \sum_{k\in\mathcal{K}}A_{ik}\hat{v}_k = \langle v,\phi^{k,\ell}_p\rangle,\label{eq:uK}
\end{eqnarray}
it is found that the same set of parameters $\{p,q,\ell\}$ also removes the noise (see Fig.~\ref{fig:reconstuction}). These results clearly demonstrate that the weak formulation with the polynomial basis chosen under appropriate $p,q,\ell$ is effectively filtering noise. {\color{black}Importantly, the expansion representation in Eq.~\eqref{eq:uK} provides a means to empirically specify the parameters $(p,q,\ell)$, which is discussed in the following subsection.}

\subsubsection{Parameter Specification}
Figure~\ref{fig:reconstuction} shows a reconstruction based on a particular choice of parameters $p,q,\ell$. In practice, we have no access to the vector field $f$ and the clean signal $u$, which prohibits comparing the filtered vector field $f_\mathcal{K}$ with $f(u)$ or the filtered signal $u_\mathcal{K}$ with $u$. 

Various techniques have been proposed to evaluate the filter quality based only on noisy observation, $v$, and the filtered output $u_\mathcal{K}$.
They include measuring the whiteness \cite{ljungbox1978}, variance ratio, smoothness ratio \cite{savitzky1964}, prediction skill compared to persistence forecast \cite{hyndman2006}.
It should be clarified that if one observes $f(v)$, then any of these metrics can be used to evaluate the quality of $f_\mathcal{K}$. In the following analysis, the focus is directed to the former case assuming that access to $f(v)$ is not necessarily available.

For parameter estimation, one can use any of these evaluation metrics depending on the nature of the data set and noise. The following score metric is considered,
\begin{equation}
J(p,q,\ell) = \frac{1}{2}\left(J_{\text{smooth}}(p,q,\ell) + J_{\text{pred}}(p,q,\ell)\right),\label{score}
\end{equation}
where
\[
J_{\mathrm{smooth}} = \left|\log\!\left(\frac{\mathrm{Var}(\Delta u_{\mathcal{K}})/\mathrm{Var}(\Delta v)}{0.2}\right)\right|, \quad\quad J_{\mathrm{pred}} = \frac{\sum_{k=2}^{N} (v_k - u_{\mathcal{K}}(t_{k-1}))^2}{\sum_{k=2}^{N} (v_k - v_{k-1})^2}.
\]
The smoothness metric $J_{\text{smooth}}$ \cite{savitzky1964} compares the variance of the smoothing local variations of the filtered signal, $\Delta u_{\mathcal{K}}(t_k)= u_{\mathcal{K}}(t_k) - u_{\mathcal{K}}(t_{k-1})$, with the variance of the noisy signal $\Delta v$. A large value of $J_{\text{smooth}}$ suggests either over- or under-smoothing, and the smoothness target is achieved when $J_{\text{smooth}} \approx 0$. The prediction metric $J_{\text{pred}}$ \cite{hyndman2006} compares the quality of the one-step persistence forecasts of the filtered solution and the noisy observation. A prediction score $J_{\text{pred}}<1$ indicates improved prediction skills and a smaller value of the prediction score is desirable.  

\begin{figure}[!htbp]
    \centering
    \includegraphics[width=\textwidth]{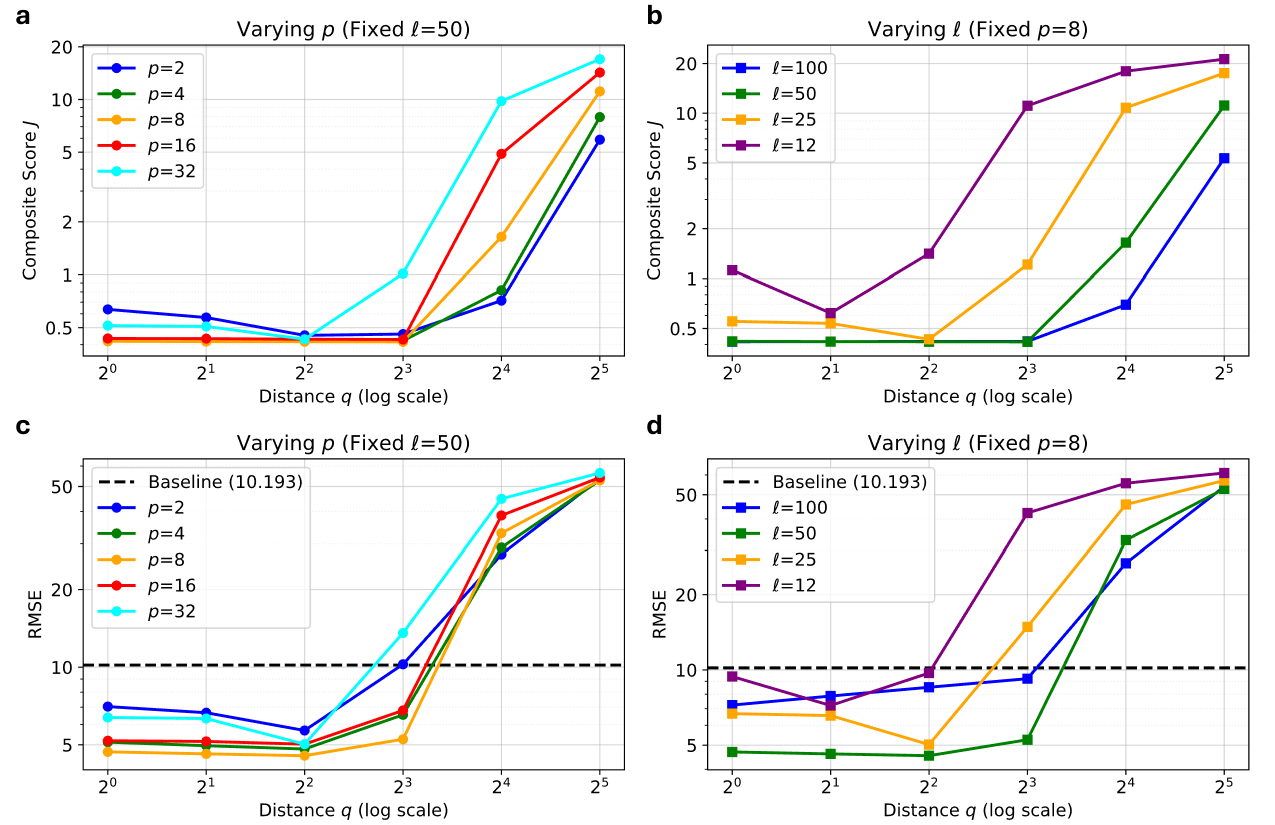}
    \caption{Score metric $J$ (top panels) and reconstruction error (RMSE) analysis (bottom panels) as functions of $q$. Panel (\textbf{a}) shows $J$ for fixed $\ell=50$. Panel (\textbf{b}) shows $J$ for fixed $p=8$. Panel
    (\textbf{c}) shows RMSE for fixed $\ell=50$. Panel
    (\textbf{d}) shows RMSE for fixed $p=8$. In the bottom panels, the black dashed line indicates the baseline observational noise. {\color{black}This analysis is conducted using the same 5\% noisy Lorenz-63 dataset as in Fig.~\ref{fig:reconstuction}.}}
    \label{fig:ablation_recon}
\end{figure}

In Fig.~\ref{fig:ablation_recon}a-b, the composite score metric $J$ is shown as a function of $q$ for various choices of $p$ and $\ell$. To observe how this metric performs, the filtered quality, measured by the root mean square error (RMSE) of the filtered signal $u_{\mathcal{K}}$ compared to the clean signal $u$, is also plotted in Fig.~\ref{fig:ablation_recon}c-d. Notice the similar trends of $J$ and RMSE as $p,q,\ell$ are varied, which suggest that $J$ is a reasonable metric for specifying these parameters. In Fig.~\ref{fig:ablation_recon}c-d, the standard deviation of the observation noise is also shown as a baseline. RMSE below the noise standard deviation clearly indicates that $v_\mathcal{K}$ is less noisy compared to $v$. It is noted that $p=8$ yields the lowest error. Excessively high orders (e.g., $p=32$) lead to an ill-conditioned Gram matrix $A$, amplifying the errors. For $\Delta t = 0.01$, $\ell=50$ yields a support of $0.5$s. Figure~\ref{fig:ablation_recon}b, d indicates that larger domain sizes (larger $\ell$) generally improve robustness against noise through averaging, provided the window size does not over-smooth the chaotic features.

The parameter $q$, which determines the distance between the centers of two closest intervals, is also inversely proportional to the number of intervals $K =|\mathcal{K}| \approx \frac{N - \ell + 1}{q}$, where $N$ is the size of the training time series. Consistent with the sensitivity analysis in Fig.~\ref{fig:L63-ablation}, reducing $q$ from 2 to 1 yields only marginal gains. Thus, $q=2$ is preferred as it reduces $K$, the number of intervals in $\mathcal{I}_{q,\ell}$ without compromising accuracy. In summary, an effective way to choose parameters $p, q, \ell $ is based on how well $v_\mathcal{K}$ induced by these parameters filters the noisy signals, which can be measured by $J$ or any other desirable metric that needs no information of the clean signal $u$. In Table~\ref{tab:sup-config-merged}, the parameters $p,q,\ell$ for weak form and \methodName{} are reported, corresponding to the numerical results in Sections~\ref{sec4} and \ref{sec5}. 

\subsection{The Proposed \methodName{}}\label{sec23}
As noted in Section~\ref{sec21}, the standard training procedure for NODE under the strong formulation is inherently sensitive to observation noise and prone to gradient instability in long-horizon rollouts. In contrast, the weak formulation suppresses high-frequency noise. However, although it enforces trajectory consistency in a weak sense across sub-intervals, the weak formulation alone does not directly assess predictive performance on time series, because it does not require explicit rollouts.

To leverage the complementary strengths of strong form and weak form supervision, these two paradigms are unified. Specifically, this paper proposes a Weak Penalty training objective for NODEs (\methodName{}). The total loss is defined as a weighted combination of Eq.~\eqref{eq:strong_loss} and Eq.~\eqref{weakloss}, balancing pointwise accuracy and integrated residual consistency:
\begin{align}\label{eq:weakpenalty}
\mathcal{L} = \mathcal{L}_{\text{weak}} + \lambda \mathcal{L}_{\text{strong}},
\end{align}
where $\lambda$ is a regularization coefficient that controls the trade-off between the two components.

A key distinction in this implementation is that $\mathcal{L}_{\text{strong}}$ is computed using significantly shorter forward-simulation trajectory window, typically using a sub-sequence of length $2 \le T \le 6$ (corresponding to $1$--$5$ rollout steps).
While the main objective of \methodName{} is to avoid the exploding gradients and instability associated with long rollouts by using shorter rollouts and enforcing long-term trajectory consistency through weak form integral constraints, the next section demonstrates that the resulting model yields stable and accurate long time predictions.

In the extreme case with single-step rollout ($T=2$), the strong loss simplifies to a derivative matching objective that avoids the need for the adjoint sensitivity method. Even when a multi-step rollout is preferred ($T \le 6$), the computational overhead remains minimal regardless of whether adjoint or direct backpropagation is used, with the latter incurring only a negligible memory cost due to the short horizon.
This design reflects a complementary role between the two loss components: The weak form loss shapes the model's global behavior via integral residuals while mitigating noise, whereas the strong form loss provides a direct, localized supervisory signal. 
By explicitly exposing the model to true temporal evolution, the strong form term reinforces accurate local dynamics and improves predictive fidelity without the cost or instability of long-horizon rollouts.

\section{Experimental Setup}\label{sec3}
The following sections compare the performance of NODE models trained using the strong form loss function in \eqref{eq:strong_loss}, the weak form loss function \eqref{weakloss}, and the Weak Penalty loss function in \eqref{eq:weakpenalty}. Whenever feasible, these methods are also compared with several baseline methods that produce competitive prediction skills, such as MP-NODE \cite{chakraborty2024divide}, VF-NODE~\cite{zhao2025accelerating}, DeepSkip \cite{mandal2025learning}, or RAFDA \cite{gottwald2021supervised}.

The following two subsections discuss the network architectures, optimization parameters, and performance evaluation metrics. 

\subsection{Network Architecture and Optimization Parameters} 
All neural networks use the GELU activation function due to its consistent performance across tasks. Shallow architectures are adopted to balance capacity and training stability.
For numerical experiments with lower-dimensional examples (Lorenz-63 and Lorenz-96 systems, which are denoted by L63 and L96, respectively, in the remainder of this paper), the vector fields are parameterized by Multilayer Perceptrons (MLPs) with 2 hidden layers, each containing 200 neurons. For a higher-dimensional example such as the Kuramoto-Sivashinsky (KS) equation, we employ a similar MLP architecture with 2 hidden layers but a larger width of 400 neurons.
Numerical integration during the training phase is performed using the \texttt{dopri5} solver with both absolute and relative tolerances set to $10^{-6}$.
Training is performed using the Adam optimizer with minibatches.
Efficient training is ensured by employing early stopping alongside a carefully tuned learning rate scheduler (specifically \texttt{ReduceLROnPlateau} with a reduction factor of $0.5$, a relative threshold of $10^{-4}$ or $10^{-7}$, and a minimum learning rate of $10^{-6}$). The patience for the scheduler and early stopping is adapted to the convergence speed of each formulation (e.g., typically 10 epochs for strong form and up to 200 epochs for weak form or \methodName{} training).
A relatively large initial learning rate of $0.002$ accelerates early convergence and is automatically decayed to enhance the performance. 

Models utilizing the weak form and \methodName{} formulations are trained for up to 20,000 epochs. In contrast, strong form models are trained for significantly fewer epochs (typically 300, depending on system complexity) due to their higher computational cost and faster convergence.
During inference, double-precision arithmetic is employed alongside the \texttt{dopri5} solver to compute all reference trajectories, ensuring sufficient accuracy for reliable evaluation.

\subsection{Evaluation Metrics}
To assess both short-term accuracy and long-term consistency, two complementary metrics are utilized: Valid Prediction Time (VPT) and Kullback–Leibler (KL) divergence~\cite{kullback1951information}.
The VPT is defined as the maximum duration (in units of Lyapunov time, $T_\Lambda = 1/\Lambda$, where $\Lambda$ is the system's maximal Lyapunov exponent), for which the normalized trajectory error remains below a certain error tolerance $\varepsilon$, 
\begin{align}
\mathrm{VPT} = \frac{1}{T_\Lambda} \max \left\{ n \Delta t \;\middle|\; E_k \leq \varepsilon \;\; \text{for all } k \leq n \right\}.\label{eq:vpt}
\end{align}
Here, the normalized error (at time step $n$) is defined as
$E_n = \sqrt{\frac{\sum_{j=1}^D \left( \hat{u}_j(n \Delta t) - u_j(n \Delta t) \right)^2}
{\sum_{j=1}^D \sigma_j^2}}$. $D$ is the dimension of the system and $\sigma_j$ is the standard deviation of the $j$-th component of the true trajectory. 

For long-term statistical consistency, the probability density function (PDF) of the surrogate model is compared to that of the reference system using the KL divergence. 
The KL divergence between a predicted distribution $P$ and a reference distribution $Q$ is formally defined as:
\begin{align}
D_{\text{KL}}(P \parallel Q) = \int P(x) \log \left( \frac{P(x)}{Q(x)} \right) dx,
\end{align}
where $P(x)$ and $Q(x)$ denote the corresponding PDFs.

For the low-dimensional L63 and L96 systems, the KDE-based KL divergence is computed dimension-wise using 1D marginal densities, and the average across all dimensions is reported. In contrast, for the spatiotemporal KS system, 1D densities lack the sensitivity required to capture the attractor's complex geometric structure. Therefore, its KL divergence is evaluated using the 2D joint PDF of the spatial derivatives ($u_x$ and $u_{xx}$), estimated via a 2D Gaussian KDE evaluated over a uniform phase-space grid.

Finally, visual comparisons of these invariant measures are presented alongside quantitative results to validate the long-term behavior, particularly in cases where the aggregate KL divergence may not fully reflect localized perceptual differences.

\section{Analysis on the Low Dimensional Lorenz-63 System}\label{sec4}
In this section, we report the detailed numerical results on the low-dimensional problem, the Lorenz-63 (L63) model. In Section~\ref{sec41}, we show representative results under an optimized choice of parameters. In Section~\ref{sec42}, we report the results of ablation analyses which allow us to learn the sensitivity of the models under perturbation of the parameters and validate the parameters specified in Section~\ref{sec22}. In Section~\ref{sec43}, we report the computational costs of \methodName{} in comparison to other NODE training methodologies.

\subsection{Lorenz-63 Analysis} \label{sec41}
The governing equations for the Lorenz-63 (L63) \cite{lorenz1963deterministic} system are:
\begin{align}
\label{eq:lorenz63}
\begin{aligned}
\frac{dx}{dt} &= \sigma (y - x), \\
\frac{dy}{dt} &= x(\rho - z) - y, \\
\frac{dz}{dt} &= xy - \beta z.
\end{aligned}
\end{align}
The well-known butterfly-like attractor can be reproduced under the 
standard parameters $\sigma = 10.0$, $\rho = 28.0$, and $\beta = \tfrac{8}{3}$. In the following numerical experiment, this system of ODEs is integrated using the RK4 scheme with a time step of $\Delta t = 0.01\mathrm{s}$ for $100\mathrm{s}$, yielding $N = 10^4$ samples. Under these parameters, the largest Lyapunov exponent is $\Lambda \approx 0.91$. To evaluate the prediction performance, we use the VPT metric in Eq. \eqref{eq:vpt} with a validation threshold of $\varepsilon = 0.3$. The weak form parameters used in this section are $p=8$, $q=2$, $\ell=50$, and $T=2$ for the strong component, as listed in Table~\ref{tab:sup-config-merged}.

\begin{figure}[!htbp]
    \centering
    \includegraphics[width=\textwidth]{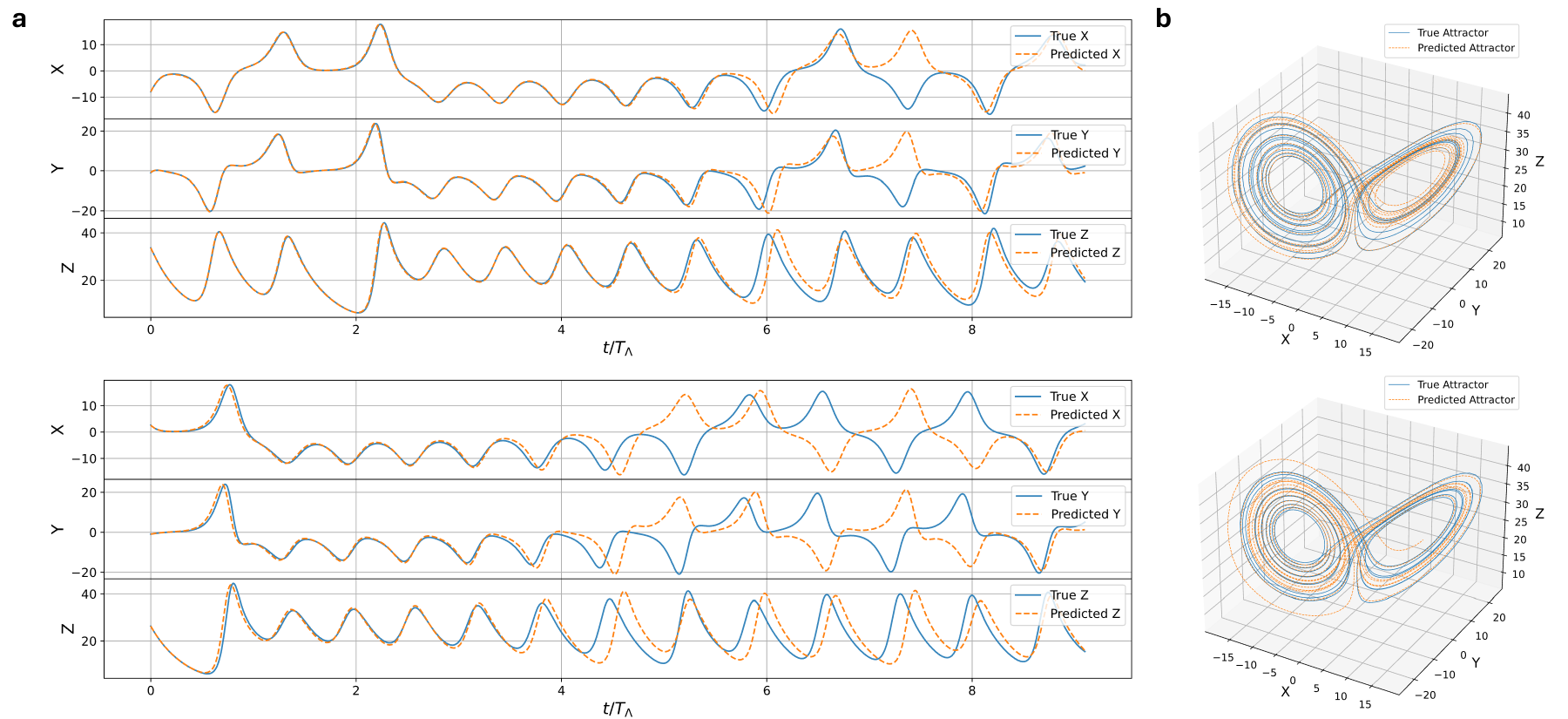}
    \caption{Forecasting of the Lorenz-63 system using \methodName{} under 5\% training data noise.
    \textbf{a}. Predictions of two unseen test initial conditions as functions of Lyapunov time, compared to the ground-truth trajectories for all three state variables. The two rows show representatives of the better (VPT = 5.77) and worse (VPT = 0.63) cases. Despite low VPT in the second case, predictions remain accurate for about 4 Lyapunov times, qualitatively speaking.
    \textbf{b}. Reconstructed phase-space attractors for the predicted (orange) and true attractor (blue).
    }
    \label{fig:lorenz63}
\end{figure}

First, $\sigma_{\text{NR}}=5\%$ data noise is applied to the observation data. The predictive skill of the trained \methodName{} model given two test initial conditions that are not in the training data is illustrated in Fig.~\ref{fig:lorenz63}, demonstrating its effectiveness in both short-time prediction and attractor recovery. Specifically, Fig.~\ref{fig:lorenz63}a shows two representative trajectories corresponding to the best and worst predictions. The resulting VPT varies between 5.77 and 0.63 Lyapunov times. Despite the lower VPT in the second case, the prediction remains visually accurate for nearly 4 Lyapunov times before diverging.
Figure~\ref{fig:lorenz63}b shows the corresponding Lorenz attractors, indicating that the \methodName{} robustly recovers the long-term statistical behavior of the system.

\begin{figure}[!htbp]
    \centering
    \includegraphics[width=\textwidth]{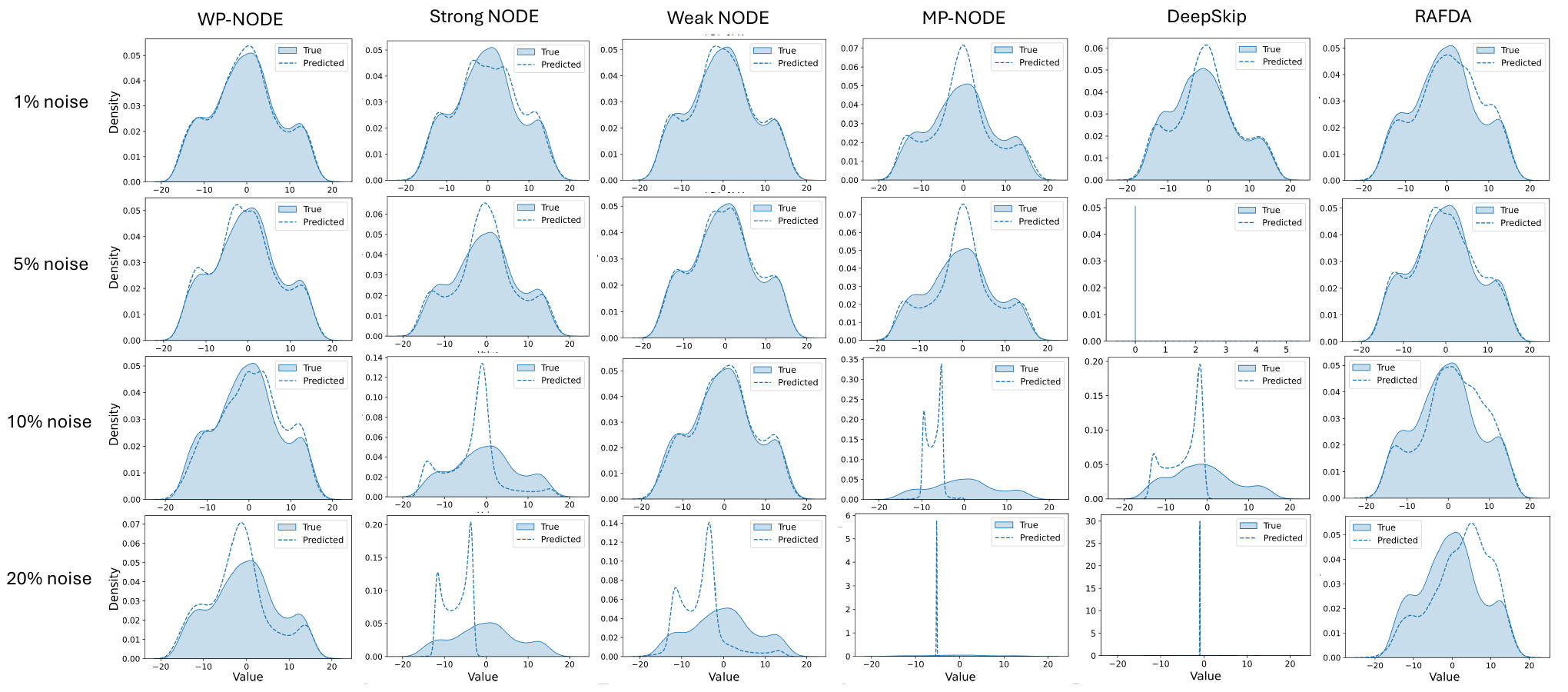}
    \caption{L63: Empirical invariant measure estimates (from $100\mathrm{s}$ prediction length). The figure shows the estimated marginal densities for $x-$component, obtained by training on noisy data of various noise levels.}
    \label{fig:lorenz63_pdf}
\end{figure}

To complement the trajectory-based evaluation, Fig.~\ref{fig:lorenz63_pdf} compares the predicted and ground-truth invariant measures using PDFs of the $x$ component, for simplicity. These PDFs represent the system’s invariant measure across representative methods and noise levels. {\color{black}Seven} methods are compared in Table~\ref{tab:cross_system_vpt_kl}, which can be grouped into the NODE family (the proposed \methodName{}, strong NODE, weak NODE, MP-NODE \cite{chakraborty2024divide}, {\color{black}and VF-NODE~\cite{zhao2025accelerating}}) and alternative approaches (DeepSkip and RAFDA). All methods perform well under very low noise (1\%). However, as noise increases to 5\%, DeepSkip performance deteriorates, indicating high sensitivity to data perturbations. {\color{black} At the same
noise level, VF-NODE's long-term rollout begins to diverge, so no invariant measure is reported (Table~\ref{tab:cross_system_vpt_kl})}. When noise is increased to 10\%, the strong, and MP-NODE estimates deteriorate. At 20\% noise, only \methodName{} and RAFDA remain highly robust, preserving a reasonable approximation of the true distribution. Quantitatively, \methodName{} 
estimates have the lowest (or close to the lowest) KL divergence (see Table~\ref{tab:cross_system_vpt_kl}).

\begin{table}[!htbp]
\centering
\setlength{\tabcolsep}{4.4pt}
\renewcommand{\ms}[2]{#1{\scriptsize$\pm$}\scalebox{0.9}{#2}}
\caption{Lorenz system evaluation (L63 and L96) under different observation noises. VPT is reported as mean $\pm$ std in Lyapunov times; KL divergence is reported as the mean value. The VPT evaluation is performed 30 times at random starting points beyond the training region. Higher is better for VPT, lower is better for KL divergence. {\color{black}'--' denotes a diverged rollout, for which the KL divergence is not reported.} RAFDA metrics for the L96 are omitted due to prohibitive computational costs (exceeding 24 hours per evaluation).}
\label{tab:cross_system_vpt_kl}
\setlength{\tabcolsep}{3.8pt}
\begin{tabular}{llcccccccccc}
\toprule
& & \multicolumn{5}{c}{VPT (Lyapunov times) $\uparrow$} & \multicolumn{5}{c}{KL divergence $\downarrow$} \\
\cmidrule(lr){3-7}\cmidrule(lr){8-12}
System & Method & 0\% & 1\% & 5\% & 10\% & 20\% & 0\% & 1\% & 5\% & 10\% & 20\% \\
\midrule
\multirow{7}{*}{\textbf{L63}}

& \textbf{\methodName{}}
& \ms{2.70}{1.03} & \ms{2.77}{1.12} & \ms{\textbf{2.90}}{1.63} & \ms{\textbf{2.23}}{1.53} & \ms{0.96}{0.59}
& 0.006 & \textbf{0.002} & 0.002 & 0.01 & \textbf{0.03} \\

& Strong NODE
& \ms{2.55}{0.97} & \ms{1.88}{0.85} & \ms{1.39}{0.84} & \ms{0.80}{0.40} & \ms{0.36}{0.16}
& \textbf{0.001} & 0.006 & 0.02 & 0.23 & 6.82 \\

& Weak NODE
& \ms{3.19}{1.55} & \ms{3.29}{1.72} & \ms{2.77}{1.77} & \ms{1.60}{0.93} & \ms{0.76}{0.59}
& 0.019 & 0.002 & {\textbf{0.001}} & {\textbf{0.003}} & 0.41 \\
& MP-NODE
& \ms{1.20}{0.71} & \ms{1.10}{0.66} & \ms{0.86}{0.46} & \ms{0.45}{0.32} & \ms{0.20}{0.12} & 0.02 & 0.04 & 0.18 & 6.67 & 11.14 \\
& {\color{black}VF-NODE}
& \ms{2.25}{1.39} & \ms{2.30}{1.38} & \ms{0.86}{0.53} & \ms{0.57}{0.28} & \ms{0.25}{0.14}
& 0.006 & 0.003 & -- & -- & -- \\
& DeepSkip
& \ms{\textbf{4.47}}{1.47} & \ms{1.54}{0.56} & \ms{0.56}{0.30} & \ms{0.34}{0.18} & \ms{0.11}{0.08} & 0.02 & 0.05 & 10.80 & 12.19 & 18.82 \\
& RAFDA
& \ms{1.51}{0.69} & \ms{\textbf{3.13}}{1.30} & \ms{2.62}{1.33} & \ms{1.91}{0.95} & \ms{\textbf{1.16}}{0.67} & 0.23 & 0.01 & 0.002 & 0.02 & 0.06 \\
\midrule

\multirow{6}{*}{\textbf{L96}}
& \textbf{\methodName{}}
& \ms{3.65}{0.64} & \ms{3.72}{0.63} & \ms{\textbf{3.24}}{0.64} & \ms{\textbf{2.75}}{0.61} & \ms{\textbf{2.08}}{0.50}
& \textbf{0.02} & \textbf{0.02} & \textbf{0.02} & \textbf{0.02} & \textbf{0.02} \\
& Strong NODE
& \ms{2.80}{0.68} & \ms{2.74}{0.58} & \ms{2.57}{0.42} & \ms{1.94}{0.38} & \ms{1.23}{0.26}
& 0.02 & 0.02 & 0.02 & 0.03 & 0.34\\
& Weak NODE
& \ms{3.66}{0.65} & \ms{3.80}{0.66} & \ms{3.19}{0.69} & \ms{2.71}{0.62} & \ms{1.99}{0.43}
& 0.02 & 0.02 & 0.02 & 0.02 & 0.02 \\
& MP-NODE
& \ms{1.46}{0.30} & \ms{1.48}{0.27} & \ms{1.35}{0.32} & \ms{1.29}{0.27} & \ms{0.93}{0.15}
& 0.02 & 0.02 & 0.04 & 0.09 & 0.77 \\
& {\color{black}VF-NODE}
& \ms{0.20}{0.02} & \ms{0.21}{0.03} & \ms{0.20}{0.03} & \ms{0.20}{0.03} & \ms{0.21}{0.03}
& -- & -- & -- & -- & -- \\
& DeepSkip
& \ms{\textbf{6.06}}{1.12} & \ms{\textbf{4.26}}{0.73} & \ms{1.64}{0.34} & \ms{1.78}{0.46} & \ms{0.63}{0.23} & 0.02 & 0.02 & 0.04 & 0.10 & 0.27 \\
\bottomrule
\end{tabular}
\end{table}

Table~\ref{tab:cross_system_vpt_kl} reports the VPT and KL divergence metrics under varying noise levels (0\%, 1\%, 5\%, 10\%, and 20\%), for all methods. As shown on the left side of the table, for the L63 system, \methodName{} consistently outperforms all other NODE variants (strong, weak, MP-NODE, {\color{black}and VF-NODE}) under noisy conditions. {\color{black}In the noiseless (0\%) regime, the pure weak NODE achieves the highest VPT {\color{black}among the NODE variants} since, as shown in Section~\ref{sec221}, the weak form loss fits the filtered representation $f_{\mathcal{K}}$, which in the absence of noise approximates the true vector field with only a small quadrature bias. However, it suffers in KL divergence because its training completely lacks forward rollout supervision, leading to distribution drift over long integration horizons. Conversely, the strong NODE captures the long-term attractor geometry best in the noiseless case (lowest KL) but accumulates numerical truncation errors during recursive rollouts, reducing its VPT. \methodName{} effectively balances these two extremes.}

When noise is low, the strong form component provides stability and accuracy, enhancing short-term prediction. As noise increases, strong form supervision {\color{black}over long trajectories (e.g., $T=25$)} becomes unreliable due to error amplification in pointwise residuals, which also severely impacts MP-NODE. In those regimes, the weak form loss plays a dominant role as it penalizes the model to the filtered data. 
Notably, even at 20\% noise, \methodName{} maintains high predictive accuracy, outperforming the weak form with a 26\% gain in VPT. 
{\color{black}While one might expect the strong penalty to degrade performance under extreme noise, its short-horizon target ($T=2$) is corrupted by the same observation noise as the weak form's local integrals, so it does not by itself point toward the true dynamics. The improvement instead reflects a trade-off between two noise-sensitive error sources: the quadrature error accumulated across the weak form's many local integral segments grows with noise, while the strong form's error, confined to a much shorter horizon, grows more slowly. Beyond a crossover that occurs between 1\% and 5\% noise (Table~\ref{tab:cross_system_vpt_kl}), the strong term helps counteract this growth, yielding both superior VPT and KL divergence for \methodName{} compared to the pure weak NODE at higher noise levels.}
Among the alternative approaches, DeepSkip, a leading baseline under clean data, deteriorates sharply with even small noise. While RAFDA demonstrates competitive robustness and high VPT for the L63 system, its current implementation \cite{gottwald2021supervised} is not  computationally feasible for application to higher-dimensional dynamics. Due to this high computational cost, we exclude RAFDA in our comparison tests on higher-dimensional examples in the remainder of this paper.

The right side of Table~\ref{tab:cross_system_vpt_kl} reports the mean KL divergence metric. While all KL metrics tend to be small under low noise, \methodName{} estimates are among the lowest values of KL, suggesting long-term statistical consistency, even at 10\% and 20\% noise levels.
We should point out that the KL divergence becomes less informative when its value is large, as it may overlook localized distributional mismatches. To complement this, Appendix~\ref{app:L63} provides visual comparisons of the invariant measure (i.e., PDFs) for all three components across noise levels. These plots show that \methodName{} consistently preserves the true distribution  while the other baseline methods degrade significantly under high noise, highlighting the superior robustness and statistical fidelity of the proposed method.

\begin{figure}[!htbp]
    \centering
    \includegraphics[width=0.9\textwidth]{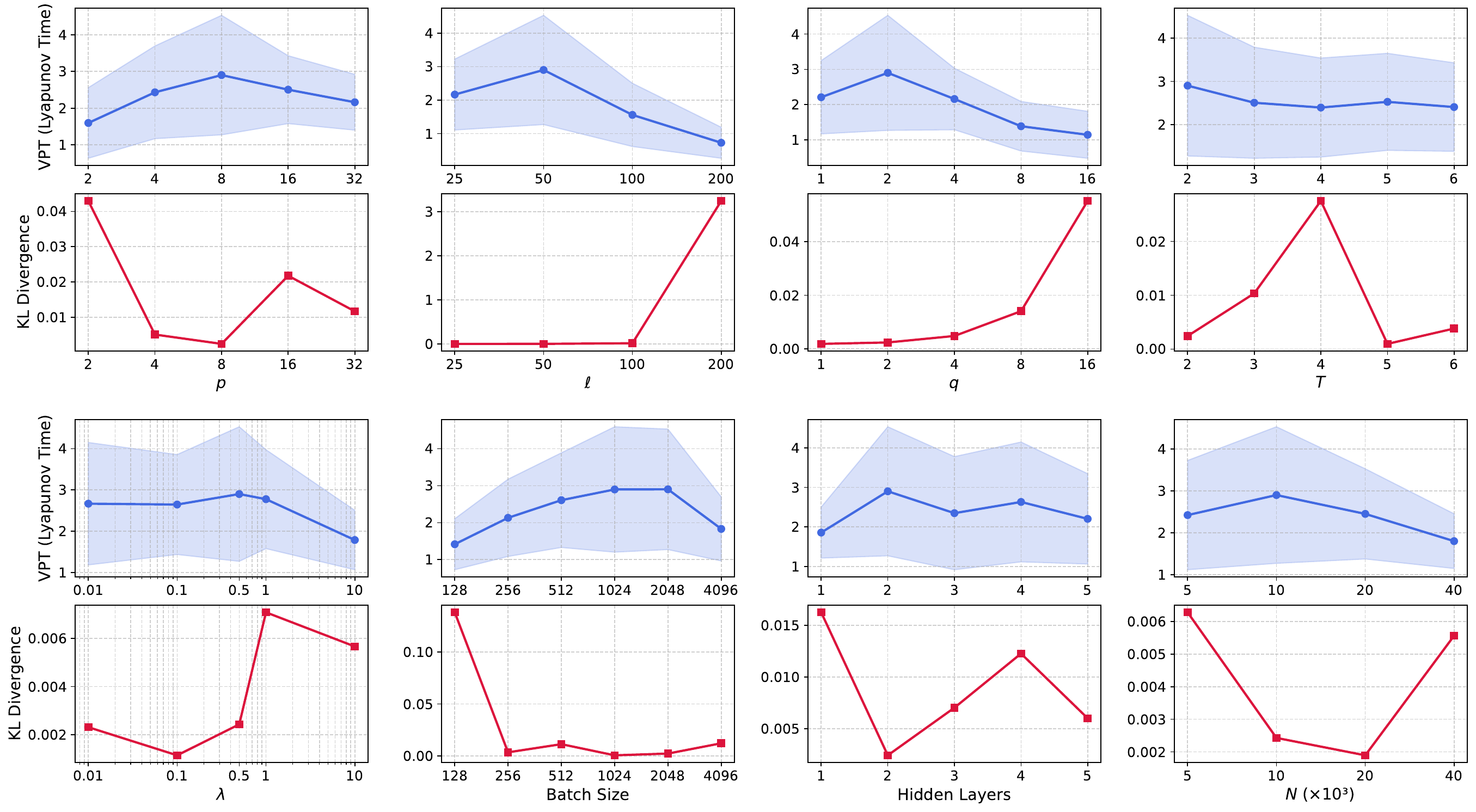}
    \caption{Ablation study of key hyperparameters for the L63 system. Evaluated parameters include $p$, $\ell$, $q$, $T$, $\lambda$, batch size, number of hidden layers, and training signal length $N$.}
    \label{fig:L63-ablation}
\end{figure}

\subsection{Hyperparameter Sensitivity}\label{sec42}
A comprehensive ablation analysis was performed to investigate the influence of key hyperparameters (Fig.~\ref{fig:L63-ablation}). 
The default configuration consists of a two-layer neural network (200 neurons per layer), and a batch size of 2048, all other parameters are specified in Table~\ref{tab:sup-config-merged}. We consider the study with noisy training data with a signal-to-noise ratio of $\sigma_{\text{NR}}=5\%$. Each parameter is varied individually while keeping the others fixed, with performance evaluated using VPT and KL divergence.

As illustrated in Fig.~\ref{fig:L63-ablation}, $p$ strongly influences the weak form integration accuracy, and values around $p = 8$ provide an optimal balance of smoothness and localization, which corresponds to the analysis in Section~\ref{sec22}. Similarly, setting $\ell=50$ and $q=2$ yields the best performance. Furthermore, increasing $T$ in \methodName{} enhances temporal consistency but reduces stability and increases computational cost (see next section for details). 
{\color{black}We should point out that with $\lambda=0.5$, while it yields the best short term prediction skill (highest VPT), it does not produce the lowest KL divergence score. We also encounter the same issue in our experiments with the KS example, where some of these parameters (especially for different values of $T$) can produce model with high VPT score but not necessary small KL divergence score or vice versa.}

Regarding network and training configurations, the batch size plays a crucial role in balancing stability and generalization. While smaller batches often lead to divergent dynamics and integration failures, excessively large batches (e.g., 4096) begin to degrade predictive performance. Consequently, a moderate batch size of 2048 is found to provide an optimal balance, stabilizing the optimization process without sacrificing generalization. Furthermore, a clear trade-off is observed regarding the network size. An architecture with two hidden layers is found to yield optimal performance; a shallower network lacks the requisite representational capacity to capture complex chaotic dynamics, whereas a deeper network tends to overfit the observational noise, resulting in degraded prediction accuracy and heightened sensitivity to input perturbations.

\subsection{Computational Cost Analysis}\label{sec43}
The training efficiency of the different formulations on the Lorenz-63 system (5\% noise) was compared using an NVIDIA A100 GPU and AMD EPYC 9555 64-core processor in Pytorch. 
All methods employ early stopping to ensure the best performance and convergence, though with different criteria adapted to their loss behaviors: the strong form uses a patience of 10 epochs (min delta $10^{-6}$), while \methodName{}, reflecting its distinct loss function, uses a patience of 200 epochs (min delta $10^{-7}$). MP-NODE is trained for a fixed maximum of 500 epochs, utilizing 10 splits and 10 rollouts per split. {\color{black}
Note that RAFDA and DeepSkip are excluded from this computational benchmarking since they are different classes of algorithms. Particularly, DeepSkip bypasses epoch-based ODE rollouts with linear regression. RAFDA employs an Ensemble Kalman filter which may need a large number of ensembles in high-dimensional problems except when additional empirical treatments such as localization and variance inflation are implemented, which require additional parameter tuning. In our numerical tests with the L96 or KS examples, we found that the current implementation exceeds 24 hours of computation time.} 

During training, it is observed that \methodName{} requires significantly more epochs to converge (1,820 epochs) compared to the strong NODE (86 epochs).
{\color{black}This difference arises because strong NODE derives gradients from long sequential rollouts, acquiring global temporal information in a single update step. In contrast, \methodName{} relies on localized integrals and short constraints, requiring more iterative steps for local gradient signals to propagate and align into a globally consistent vector field. However,} \methodName{} ($T=2$) requires only $\sim$0.077~s per epoch, which is over five times faster than the strong formulation ($\sim$0.407~s/epoch).
This per-epoch speed highlights a fundamental structural advantage: while strong NODE relies on strictly sequential ODE solvers over long trajectories, the weak component of \methodName{} uses highly parallelizable numerical quadrature, and its strong component operates on a minimal horizon ($T=2$), rendering ODE solver costs negligible. Meanwhile, MP-NODE incurs a massive computational overhead, taking $\sim$~1.488~s/epoch. {\color{black}VF-NODE also avoids ODE integration and is therefore inexpensive per epoch ($\sim$0.007~s/epoch), but its global integrals span the entire training record and cannot be mini-batched.}

Regarding the total wall-clock training time, the strong NODE converges the fastest ($35$~s) due to its extremely low epoch count, followed by the baseline weak NODE ($81$~s).
Although MP-NODE requires fewer epochs to converge compared to \methodName{}, its per-epoch cost is exceptionally high due to the heavy computation through the ODE solver across multiple sequence splits and calculating complex penalty gradients. Consequently, its total training time reaches $744$~s. 
{\color{black}For \methodName{}, the longer total training time ($140$~s) compared to the pure strong or weak NODE represents a necessary computational trade-off required to robustly propagate local constraints into a global vector field without numerical instability.} 
It is noted that the exact total time depends on the specific early stopping criteria and the level of observation noise, which can influence convergence dynamics. 
{\color{black}Nevertheless, despite the significantly higher epoch count, the total training time of \methodName{} remains within a practical range.} Detailed metrics are illustrated in Table \ref{tab:computation}.

\begin{table}[!htbp]
    \centering
    \caption{Computational cost comparison of different NODE formulations on the L63 system (5\% observation noise).\label{tab:computation}}
    \begin{tabular}{lccc}
        \toprule
        \textbf{Method} & \textbf{Per-epoch Time (s)} & \textbf{Epochs} & \textbf{Total Time (s)} \\
        \midrule
        \textbf{\methodName{}} ($T=2$) & 0.077 & 1,820 & 140 \\
        Strong NODE & 0.407 & 86 & 35 \\
        Weak NODE & 0.048 & 1,673 & 81 \\
        MP-NODE & 1.488 & 500 & 744 \\
        {\color{black}VF-NODE} & 0.007 & 20,000 & 131\\
        \bottomrule
    \end{tabular}
\end{table}

\section{Results on Higher Dimensional Systems}\label{sec5}
In this section, we report numerical results on two synthetic higher-dimensional problems, L96 and KS {\color{black}(both integrated using the dopri5 solver)}, in Sections~\ref{sec51} and \ref{sec52}, respectively.  {\color{black}Following the noise model defined in Eq.~\eqref{noisyobs},
observation noise is added to the training data at five noise levels:
$\sigma_{\text{NR}} \in \{0\%, 1\%, 5\%, 10\%, 20\%\}$.} Additionally, we show results on the atmospheric Reanalysis (ERA5) dataset {\color{black}(integrated using the Euler method)} in Section~\ref{sec53}.

\subsection{Lorenz-96}\label{sec51}
To evaluate performance on high-dimensional chaotic dynamics, \methodName{} is tested on the 40D L96~\cite{lorenz1996predictability} model, 
\begin{align}
\label{eq:lorenz96}
\frac{dx_i}{dt} = (x_{i+1} - x_{i-2}) x_{i-1} - x_i + F, \quad i=1,\dots,d,
\end{align}
with cyclic boundary conditions $x_{i\pm kd} = x_i$ and forcing $F=10${\color{black}, where the system dimension is $d=40$}. The system is integrated with step size $\Delta t = 0.01$~s for $1000$~s ($N=10^5$ steps). The VPT is measured with $\varepsilon = 0.5$. This setup yields a leading Lyapunov exponent of $\Lambda \approx 1.68$~\cite{mandal2025learning}.

As shown in Table~\ref{tab:cross_system_vpt_kl}, a clear performance gap emerges within the NODE family for most experiments. The strong form methods (strong NODE and MP-NODE) exhibit significantly lower VPT scores and higher KL divergence, confirming that pointwise supervision becomes a source of instability when high-dimensional chaos is coupled with observational noise. {\color{black}While DeepSkip performs competitively under noiseless or very low noise regimes (0\% and 1\%) with strong VPT and KL scores, it lacks the same robustness at higher noise levels.}
{\color{black}VF-NODE fails outright on this higher-dimensional system: its VPT stays near $0.2$ Lyapunov times at every noise level, including the noiseless case, and its long-term rollout diverges throughout, so no KL divergence is reported (Table~\ref{tab:cross_system_vpt_kl}). It is therefore not extended to the KS system.}
In contrast, \methodName{} and the weak NODE achieve the highest and nearly identical performance across all noise levels. This suggests that for the L96 system, the robust integral constraints of the weak formulation provide the primary drive for learning the dynamics. Notably, \methodName{} does not suffer from the performance degradation seen in other strong form variants. {\color{black}As observed in the L63 system, its short-horizon strong penalty is itself affected by the same observation noise, so it does not correct the weak formulation toward the true dynamics; its error grows more slowly with noise than the quadrature error accumulated across the weak form's local integral segments, so it helps counteract the weak formulation's increasing instability under noise.}

\begin{figure}[!htbp]
    \centering
    \includegraphics[width=0.9\textwidth]{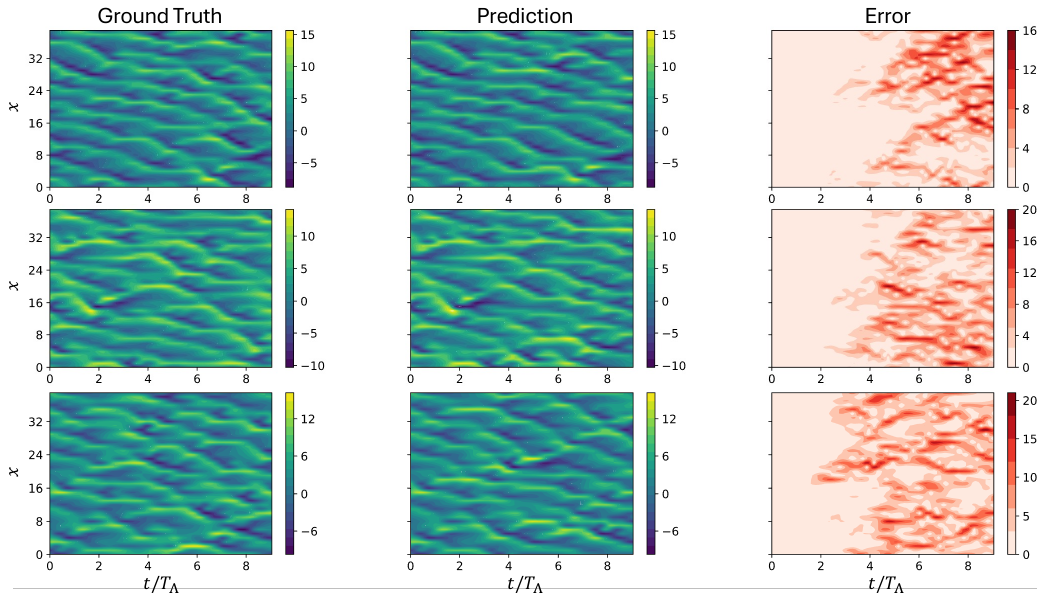}
    \caption{Forecasting of the Lorenz–96 system using \methodName{} under 5\% training data noise. The three rows correspond to three different initial conditions for forecasting, representing the best, average, and worst cases. From top to bottom, the VPT scores are 4.29, 3.31, and 1.97.}
    \label{fig:L96-dynamics}
\end{figure}

\begin{figure}[!htbp]
    \centering
    \includegraphics[width=\textwidth]{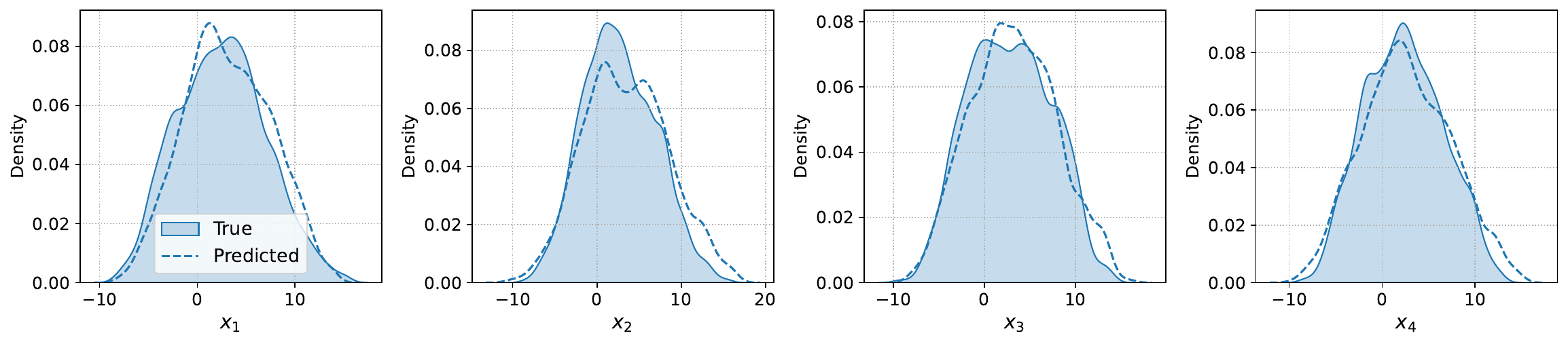}
    \caption{Invariant measures for first four components ($x_1, x_{2}, x_{3}, x_{4}$) of the learned 40D Lorenz–96 system using \methodName{} under 5\% training noise. The model consistently recovers the reference distribution across the spatial domain. The corresponding KL divergence values for these marginal densities are $0.016$, $0.042$, $0.028$, and $0.010$, respectively.}
    \label{fig:L96-pdf-4dims}
\end{figure}

Figure~\ref{fig:L96-dynamics} illustrates {\color{black}representative forecasting performances (best, average, and worst cases)} of \methodName{} under 5\% noise for three initial conditions, where the model maintains spatiotemporal coherence over multiple Lyapunov times. {\color{black}In terms of statistical consistency (as measured by KL divergence), most evaluated models capture the distributions reasonably well, likely because the marginal densities of the L96 system resemble a Gaussian profile.} In {\color{black}Fig.~\ref{fig:L96-pdf-4dims}, we show the invariant measures for the first four components of the dynamics (the other components look very similar), demonstrating that while \methodName{} can accurately recover the support and the peak of these marginal densities, there is still room to improve the estimation of the detailed features of the marginal density in this high-dimensional problem.}

\subsection{Kuramoto--Sivashinsky} \label{sec52}
\begin{table}[!htbp]
\centering
\setlength{\tabcolsep}{4.4pt}
\renewcommand{\ms}[2]{#1{\scriptsize$\pm$}\scalebox{0.9}{#2}}
\caption{KS evaluation under different observation noises. VPT is reported as mean $\pm$ std in Lyapunov times; KL divergence is reported as the mean value. $T$ denotes the trajectory window size.
{\color{black}VF-NODE is not included, see Section~\ref{sec51}}. }
\label{tab:ks_vpt_kl}
\setlength{\tabcolsep}{3.8pt}
\begin{tabular}{lcccccccccc}
\toprule
& \multicolumn{5}{c}{VPT (Lyapunov times) $\uparrow$} & \multicolumn{5}{c}{KL divergence $\downarrow$} \\
\cmidrule(lr){2-6}\cmidrule(lr){7-11}
Method & 0\% & 1\% & 5\% & 10\% & 20\% & 0\% & 1\% & 5\% & 10\% & 20\% \\
\midrule
\textbf{\methodName{}}
& \ms{3.38}{0.79} & \ms{3.71}{1.17} & \ms{\textbf{3.31}}{0.95} & \ms{\textbf{2.87}}{0.97} & \ms{\textbf{1.53}}{0.58}
& 0.93 & 0.12 & \textbf{0.008} & 0.02 & 0.10 \\
Strong NODE ($T=50$)
& \ms{2.78}{0.84} & \ms{2.67}{0.76} & \ms{2.89}{0.75} & \ms{2.12}{0.54} & \ms{1.27}{0.41}
& 0.16 & 0.21 & 0.18 & 0.15 & 0.13 \\
Strong NODE ($T=25$)
& \ms{2.61}{0.74} & \ms{2.49}{0.42} & \ms{2.45}{0.76} & \ms{1.88}{0.60} & \ms{1.32}{0.40}
& 0.45 & 0.08 & 0.01 & \textbf{0.01} & \textbf{0.02} \\
Weak NODE
& \ms{3.49}{0.92} & \ms{\textbf{3.82}}{0.66} & \ms{1.97}{0.46} & \ms{2.42}{0.89} & \ms{0.44}{0.11}
& 1.78 & 1.57 & 1.90 & 1.90 & 1.90 \\
MP-NODE
& \ms{1.63}{0.55} & \ms{2.13}{0.56} & \ms{2.37}{0.72} & \ms{1.57}{0.46} & \ms{1.14}{0.32} & \textbf{0.13} & 0.10 & 0.009 & 0.10 & 0.46 \\
DeepSkip
& \ms{\textbf{4.26}}{1.46} & \ms{2.51}{0.93} & \ms{1.16}{0.52} & \ms{0.76}{0.49} & \ms{0.45}{0.33} 
& 0.03 & \textbf{0.01} & 0.034 & 0.09 & 1.71 \\
\bottomrule
\end{tabular}
\end{table}

The KS equation \cite{kuramoto1976persistent, sivashinsky1977nonlinear} is a canonical benchmark for spatiotemporal chaos in nonlinear PDEs. It describes the evolution of a field $u(x,t)$ governed by
\begin{align}
\label{eq:ks}
\frac{\partial u}{\partial t} 
+ u \frac{\partial u}{\partial x} 
+ \frac{\partial^2 u}{\partial x^2} 
+ \nu\frac{\partial^4 u}{\partial x^4} = 0,
\end{align}
where chaotic behavior emerges from the interaction between nonlinearity, instability, and dissipation. In this standard chaotic regime, the hyperviscosity coefficient is set to $\nu=1$. The system is defined on a periodic spatial domain of length 22, which provides sufficient extent to sustain chaotic fluctuations. For numerical integration and data-driven modeling, this physical domain is uniformly discretized into a $64$-dimensional state vector.

{\color{black}To adapt this PDE to the NODE framework,} the system is discretized using 64 Fourier modes on a periodic domain of length $22$, and integrated for $N = 10^5$ steps with a time step of $\Delta t = 0.25$, yielding a $25{,}000$~s trajectory. The VPT is measured with $\varepsilon = 0.5$. In this regime, the largest Lyapunov exponent is approximately $\Lambda \approx 0.05$~\cite{fan2020long}.

\begin{figure}[!htbp]
    \centering
    \includegraphics[width=\textwidth]{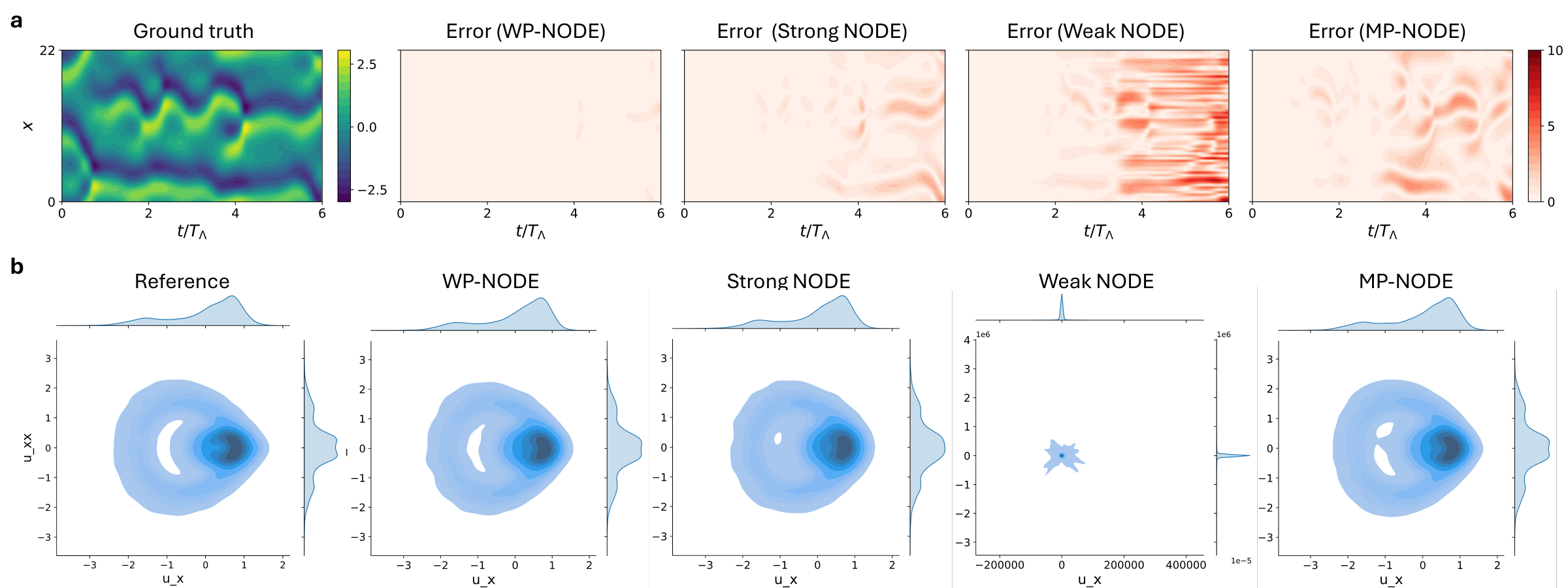}
    \caption{Performance comparison of \methodName{} and baselines models on the KS system, under 5\% data noise. 
    \textbf{a}. Short-time prediction comparison. {\color{black}We note that strong NODE corresponds to $T=50$. } \methodName{} achieves the lowest error (highest VPT). 
    \textbf{b}. Joint probability density for the KS system under 5\% observation noise. \methodName{}, strong NODE {\color{black}($T=25$)}, and MP-NODE closely reproduce the reference distribution, with \methodName{} showing improved alignment in the core region. The weak NODE, however, exhibits notable distortion in the invariant measure.}
    \label{fig:KS-dynamics-pdf}
\end{figure}

Table~\ref{tab:ks_vpt_kl} reports robustness under different signal-to-noise ratios, {\color{black}evaluating both short-term tracking (VPT) and long-term statistical consistency (KL divergence)}.
{\color{black}As noted in the previous discussion, we observe that a parameter value of $T$ in the strong NODE that produces more accurate short-time prediction does not necessarily produce accurate long-time prediction and vice versa.} In Table~\ref{tab:ks_vpt_kl}, we report results with $T=25$ and $T=50$, where $T=25$ is retained for consistency with previous experiments, while the extended window $T=50$ is included to observe the model's behavior under a longer rollout horizon.

In contrast, \methodName{} successfully overcomes this trade-off. For short-term prediction, it maintains stable performance under perturbations, achieving a VPT of 2.87 at 10\% noise, while the strong NODE ($T=50$) baseline drops to {\color{black}2.12}.
At higher noise levels, \methodName{} continues to outperform other baselines {\color{black}in short-term prediction}, highlighting its superior robustness. {\color{black}Simultaneously, \methodName{} achieves highly competitive KL divergence scores across noise levels, avoiding the statistical distortion seen in purely strong form training.}
Interestingly, the weak NODE baseline shows erratic behavior: its VPT at 5\% noise (1.97) is even lower than at 10\% (2.42), indicating instability in dynamics reconstruction, consistent with prior findings~\cite{zhao2025accelerating}.

{\color{black}These quantitative observations are further supported by visual comparisons.} 
Figure~\ref{fig:KS-dynamics-pdf}a presents the short-time prediction skill of \methodName{} and related baselines on the KS system, where these models are trained with time series corrupted by 5\% noise.  It is noted that \methodName{} faithfully reproduces the dominant spatiotemporal structures, achieving a mean VPT of 3.31 (see Table~\ref{tab:ks_vpt_kl}). In contrast, {\color{black}the strong NODE (using $T=50$)} and weak NODEs exhibit increasing phase drift and amplitude decay, with shorter predictive windows of {\color{black}2.89} and 1.97, respectively. 
Figure~\ref{fig:KS-dynamics-pdf}b highlights \methodName{}’s ability to recover accurate long-term statistics via joint probability density plots. Under 5\% observation noise, both \methodName{} and the strong NODE {\color{black}(here, we show the better estimate corresponding to $T=25$)} approximate the reference invariant distribution, but only \methodName{} captures both the core and tail behaviors with high fidelity. In contrast, the weak form baseline fails to reproduce the full statistical structure, indicating degraded long-horizon dynamics. 
{\color{black}Regarding the non-monotonic KL divergence trend of \methodName{}, this stems from the noiseless (0\%) case, where the weak-form target contains only the quadrature bias, which the model fits along with the dynamics; moderate noise acts as an implicit regularizer that mitigates this effect. Furthermore, the superior VPT of \methodName{} at 20\% noise compared to the pure weak NODE is consistent with the L63 case: the quadrature error accumulated across the weak form's local integral segments grows faster with noise than the error of the short-horizon strong component, so the strong term counteracts the weak formulation's increasing instability under extreme noise.}

Overall, these results underscore a key strength of \methodName{}: {\color{black}while purely strong form models often sacrifice short-term tracking for long-term statistics (or vice versa), \methodName{} significantly extends predictive horizons under noise without compromising the general statistical structure of the chaotic system. This robust result is crucial for reliable spatiotemporal modeling. For a comprehensive visual comparison of these invariant measures across all evaluated noise levels and baseline variants, we refer the reader to Appendix~\ref{app:ks}.}

\begin{figure}[!htbp]
    \centering
    \includegraphics[width=\textwidth]{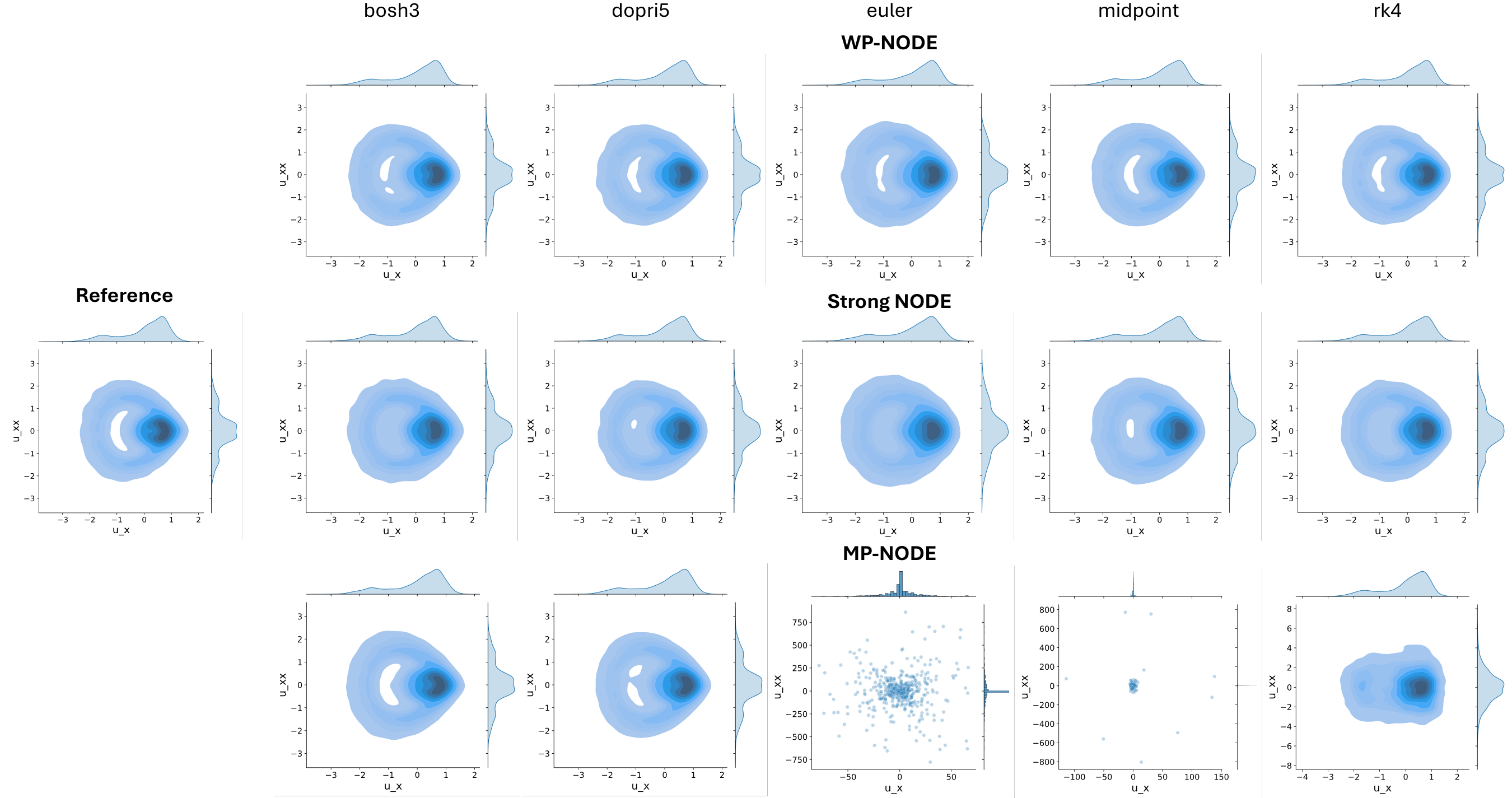}
    \caption{Comparison of the long-term behavior of the KS system under 5\% observation noise, visualized via the 2D invariant measure. 
    \methodName{} (top row) yields highly consistent density estimates across all ODE solvers, outperforming strong NODE (middle row) and MP-NODE (bottom row). Note that MP-NODE diverged when integrated with the \texttt{euler} and \texttt{midpoint} solvers; these diverging trajectories were truncated early, and the failed cases are depicted as truncated scatter plots rather than KDE contours.}
    \label{fig:KS_solver}
\end{figure}

\paragraph{{\color{black}Robustness to numerical solvers.}}
To further examine the sensitivity of the NODE model to ODE integrators, the models (originally trained using the \texttt{dopri5} solver) were integrated under different numerical integration schemes, including \texttt{bosh3}, \texttt{euler}, \texttt{midpoint}, and \texttt{rk4}, for a $400$~s prediction (approximately 20 Lyapunov times) with 5\% observation noise. 
As shown in Fig.~\ref{fig:KS_solver}, the strong NODE, despite being trained with 25-step rollouts, produces varying invariant measures across solvers, highlighting its sensitivity to the discretization scheme used during training.
{\color{black}Furthermore, MP-NODE, utilizing 5 splits and 10 rollouts per split in the KS system, struggles to generalize across integration schemes. It severely diverges when evaluated with lower-order fixed-step solvers such as \texttt{euler} and \texttt{midpoint}, resulting in numerical overflow. Consequently, these exploding trajectories were truncated early, and the failed cases are visualized as truncated scatter plots in Fig.~\ref{fig:KS_solver}.}
The weak NODE, lacking any rollout-based supervision, fails to capture long-term dynamics and becomes unstable across all solvers. Thus, the joint probability density estimates from various ODE solvers are omitted. 
{\color{black}In contrast, \methodName{} consistently reproduces the reference density across all tested solvers.
This robustness indicates that it successfully captures the true continuous-time dynamics rather than overfitting to a specific discretization scheme, effectively overcoming the solver sensitivity and long-term instability that plague purely strong or weak NODEs.}

\begin{figure}[!htbp]
    \centering
    \includegraphics[width=\textwidth]{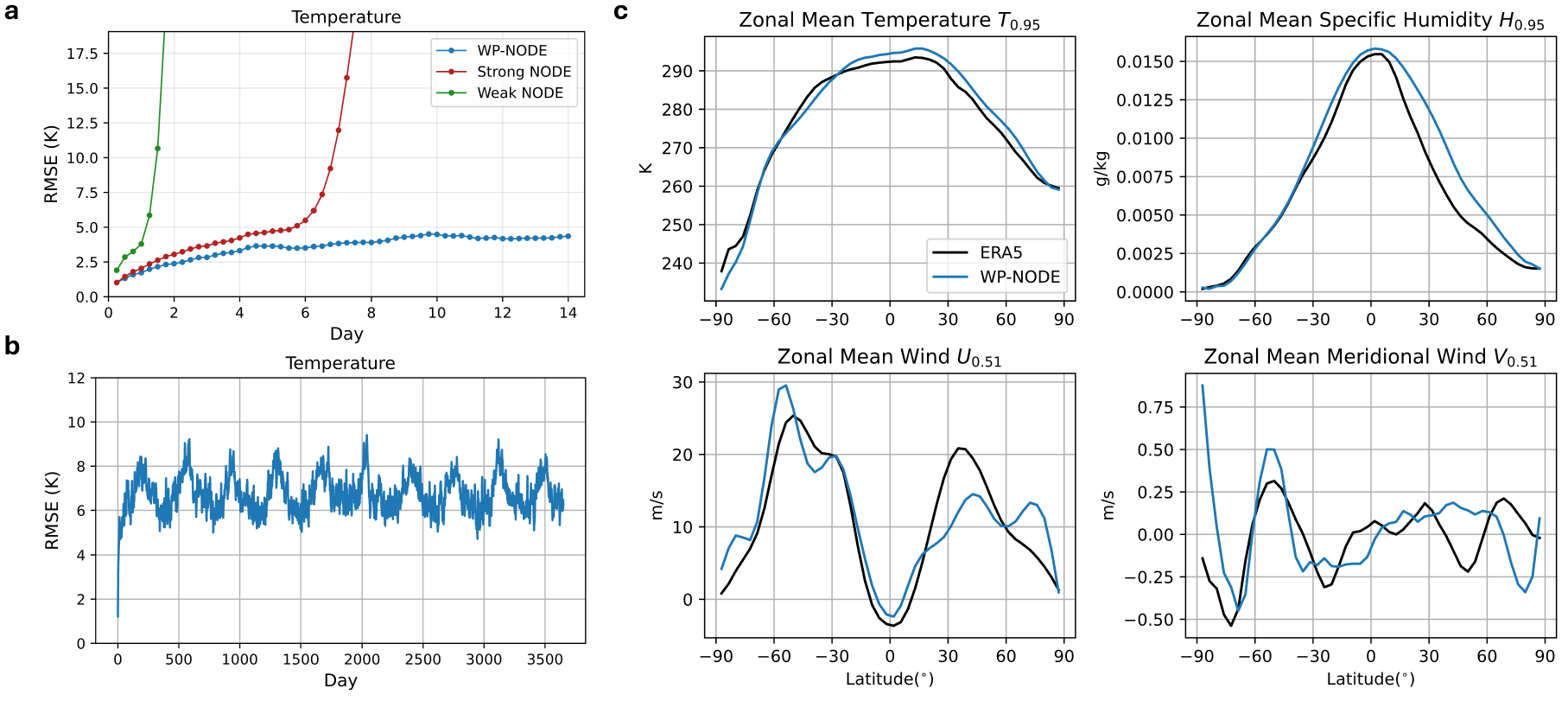}
    \caption{NODE predictions performance on real-world ERA5 dataset. 
    \textbf{a}. Short-term (14-day) temperature prediction errors. \methodName{} achieves the lowest error. 
    \textbf{b}. Long-term (10-year) temperature prediction errors. \methodName{} remains stable. 
    \textbf{c}. Climatological statistical prediction for 10 years of \methodName{} across the four ERA5 prognostic variables. 
    } 
   \label{fig:ERA5}
\end{figure}

\subsection{ERA5 Atmospheric Reanalysis Dataset}\label{sec53}
To evaluate \methodName{} in a high-dimensional, noisy, and strongly coupled real-world setting, experiments are conducted on the ERA5 atmospheric reanalysis dataset with a temporal resolution of 6 hours, {\color{black} spanning approximately 11.3 years}~\cite{hersbach2020era5}. The data are regridded to a T30 Gaussian grid and vertically interpolated onto $\sigma$-levels following the preprocessing of~\cite{arcomano2022hybrid}, yielding input tensors of four prognostic variables, including near-surface temperature and specific humidity at $\sigma_{0.95}$, and mid-atmospheric zonal ($u$) and meridional ($v$) winds at $\sigma_{0.51}$. The horizontal resolution of each variable consists of 48 latitude times 96 longitudinal data points. Total Incoming Solar Radiation (TISR) is added as a temporal embedding to capture external radiative forcing~\cite{guan2024lucie}.

The ERA5 dataset, comprising a total of 16,496 samples, was subjected to min-max normalization and partitioned chronologically. The training set contains 11,576 samples, spanning approximately 7.9 years. We evaluate the short-term prediction skill using initial conditions drawn exclusively from the unseen testing set (4,920 samples). To assess the long-term statistical consistency, we compare the climatological averages of the ground truth against those captured by a 10-year model rollout (14,600 steps). Since the testing set alone is shorter than 10 years, this rollout is initialized from an earlier state to span the continuous final 10 years of the entire dataset.

To effectively manage these high-dimensional spatial dependencies, we employ a structured encoder-processor-decoder architecture rather than a standard feed-forward network. This design operates in a compressed latent space, significantly reducing computational costs while maintaining high expressivity. 
The encoder compresses the input state and conditioning variables, which are then processed by a dilated CNN to capture multi-scale patterns. Finally, the decoder maps these latent features back to the physical space to output the predicted rate of change.

To correctly model the Earth's spherical shape, all convolutional operations use cylindrical boundary conditions: circular padding for longitude (to connect East and West) and replicate padding for latitude (to handle the poles). 
Specifically, the encoder and decoder consist of two $3 \times 3$ convolutional layers (stride 1, 64 channels, GELU activations). The dilated processor utilizes residual blocks comprising 12 convolutional layers (maintaining the 64-channel width) with a dilation schedule of $[1, 2, 3, 4, 8, 16, 32, 16, 8, 4, 3, 2]$. 
No spatial downsampling is performed within the network to preserve the resolution of the physical grid. The temporal integration is performed using the Euler method.
While the forward Euler method has poor stability properties, it is adopted here because the large CNN architecture makes higher-order adaptive solvers computationally prohibitive. Furthermore, the inherent instability of this scheme effectively highlights the robustness of \methodName{}, which maintains stable long-term rollouts where standard baselines fail.
Training is optimized using AdamW with automatic mixed precision and gradient clipping to ensure efficiency and stability.

Since solar forcing varies with time and location, the proposed NODE model accepts conditioning information as an additional input. The discrete update reads:
\begin{align}
u_{n+1}= u_n + f(u_n, x_n, t_n;\theta)\Delta t,
\end{align}
where $u_n$ represents the atmospheric state, $x_n$ is the time-dependent input forcing (i.e., solar radiation), $\theta$ denotes the learnable parameters, and $\Delta t$ is the integration time step. The time variable is non-dimensionalized by the data sampling interval (6 hours). Therefore, the integration step size is set to $\Delta t = 1$ in the discrete model, representing one step of forward prediction corresponding to a 6-hour physical duration.
During training, the trajectory lengths are kept short to maintain computational efficiency. The strong NODE uses $T=6$, whereas the weak NODE adopts $\ell=8$, balancing the need for temporal coverage with memory limitations. The distance parameter $q$ is set to 1 due to the limited number of samples and small window size.

{\color{black}To ensure a fair evaluation of the training strategies,} the comparison is restricted to NODE-based baselines {\color{black}utilizing the identical latent CNN backbone. }
Notably, the MP-NODE baseline is excluded from this evaluation, as its multi-step penalty formulation incurs prohibitive computational costs and unfeasible training times (as discussed in Sec.~\ref{sec43}).

Following model training, both the short- and long-term predictive capacities of \methodName{} are assessed and compared against the baseline weak and strong NODEs.
In the short-term forecasting regime (Fig.~\ref{fig:ERA5}a, 14 days), both purely weak and strong NODEs accumulate errors rapidly and exhibit numerical instability. In contrast, \methodName{} effectively mitigates this error growth, consistently achieving the lowest RMSE.
{\color{black}This improvement stems from the synergistic interaction between the spatial filtering of the CNN encoder and the temporal filtering of the weak-form integral constraints. The encoder suppresses spatial noise through local convolutions and latent-space compression, but this alone cannot prevent error accumulation along the time dimension. This is evidenced by the divergence of the strong NODE baseline, which shares the identical encoder. The weak penalty directly constrains this temporal error growth.}
Remarkably, this stability extends to the 10-year horizon (Fig.~\ref{fig:ERA5}b). While both baseline NODEs diverge completely, \methodName{} remains bounded, demonstrating its capability to capture the system's long-term dynamics without unnatural energy accumulation.
Figure~\ref{fig:ERA5}c further evaluates the predicted climatological statistics. 
For each variable, the temporal average over the 10-year testing period is computed. 
The zonal mean at each latitude is then obtained by averaging the time-averaged field along the longitude direction.
We observe that \methodName{} produces accurate predictions of both the zonal mean temperature and specific humidity. 
Although the predictions for the average zonal and meridional winds exhibit slight deviations, \methodName{} successfully captures their overall latitudinal trends. This behavior is physically consistent, as mid-atmospheric wind fields are inherently more chaotic and dominated by high-frequency transient eddies compared to large-scale thermodynamic structures.
Detailed configurations and additional performance comparisons, including extended 1-year predictions for other prognostic variables, are provided in Appendix~\ref{app:era5}.

Overall, the ERA5 experiment highlights the robustness of this hybrid training strategy in stabilizing NODE-based learning. It demonstrates that, even with a relatively simple CNN-based architecture, \methodName{} delivers superior stability, accuracy, and long-term predictive fidelity on real-world climate data, successfully operating in regimes where purely weak and strong NODEs deteriorate.

\section{Conclusion and Discussion}
In this paper, we verified that the weak formulation proposed by Bortz et al. \cite{bortz2024weak} is analogous to filtering when its hyperparameters are chosen appropriately. Building on this, we used classical filtering metrics to parameterize a weak loss function for training dynamical systems. We incorporated the weak loss as a penalty in the training of a Neural ODE, creating the Weak Penalty NODE (\methodName{}). 

Our numerical results show that this Weak Penalty stabilizes and improves the predictive skill of standard Neural ODEs for chaotic systems or noisy data.
Across three benchmark chaotic systems (L63, L96, and KS), \methodName{} consistently delivers accurate short-term forecasts and preserves long-term statistical properties, even with significant observational noise. {\color{black}Importantly, the resulting NODE model produces accurate and robust long-term statistical predictions under 
various ODE solvers that are different from the one used in training.} When applied to the high-dimensional ERA5 climate reanalysis dataset, \methodName{} scales effectively, maintaining stable and accurate long-term forecasts where other Neural ODE methods quickly fail.

{\color{black}Furthermore, the short-horizon strong component of \methodName{} keeps the per-epoch cost low, since backpropagation through the ODE solver is confined to at most a few steps and the weak component requires no ODE integration at all. While \methodName{} requires more training epochs, the total wall-clock training time remains within a practical range, reflecting a necessary trade-off between local gradient propagation and global consistency.}
{\color{black}Regarding data sampling, while our current implementation assumes uniform spacing, the weak form quadrature can be readily adapted to non-uniform grids. Furthermore, accurate integral approximation requires sufficiently dense observations, which precludes the original NODE's ability to handle large temporal gaps. We view this density requirement as a physical necessity for modeling chaotic systems; attempting to bridge sparse chaotic data via interpolation (e.g., \cite{zhao2025accelerating}) introduces structural bias and degrades the true phase-space geometry.}
Overall, this work demonstrates that the weak formulation offers a simple yet powerful method for fitting models to filtered data, and we plan to explore its application to other model classes beyond Neural ODEs and compare with other denoising algorithms. 


\section*{Acknowledgments}
This research used resources of the National Energy Research Scientific Computing Center (NERSC), a U.S. Department of Energy Office of Science User Facility, and the Institute for Computational and Data Sciences (ICDS) at Pennsylvania State University, and the University Research Computing (URC) at the University of North Carolina at Charlotte on the Orion cluster.
R.M. acknowledges support from the U.S. Army Research Office (ARO) Young Investigator Program under the Multiscale Modeling of Complex Systems, PM - Rob Martin.
The research of J.H. was partially supported by the NSF Grant DMS-2505605.

\appendix
\section{Lorenz 63 Analysis}\label{app:L63}
To complement the quantitative KL divergence results presented in Table~\ref{tab:cross_system_vpt_kl} of the article, this note provides a detailed visual comparison of the predicted and ground-truth PDFs for the L63 system under different modeling approaches. These comparisons offer a more intuitive understanding of how well each method captures the system's invariant measure, especially under various noise levels.

\begin{figure}[!htbp]
    \centering
    \includegraphics[width=\textwidth]{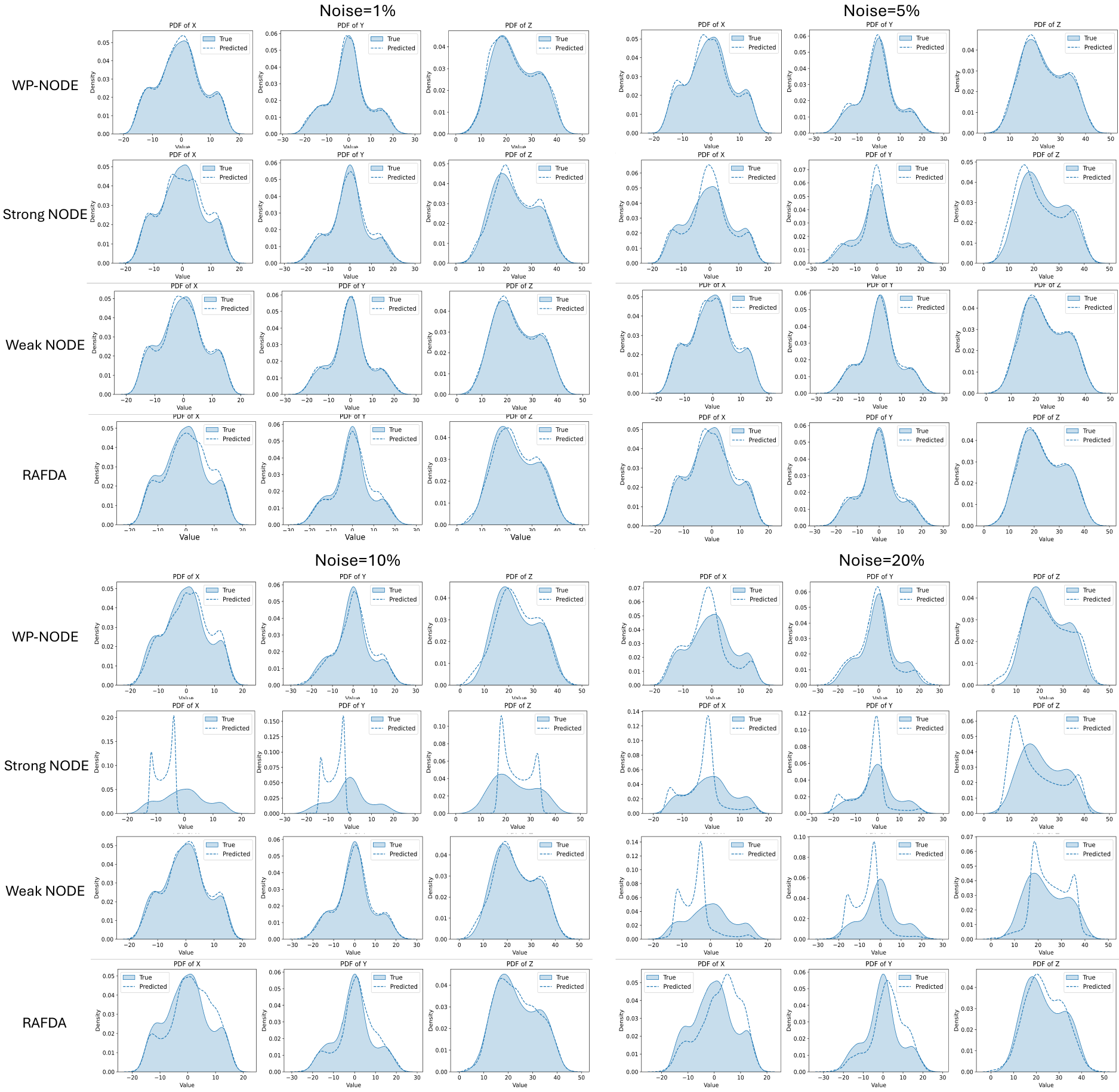}
    \caption{Invariant measure comparisons of the learned Lorenz–63 system across different methods under varying noise conditions. MP-NODE and DeepSkip are omitted due to severe distribution collapse under moderate to high noise.}
    \label{fig:L63-pdf-sup}
\end{figure}

{\color{black}Notably, all evaluations are performed on trajectories generated beyond the training time interval, starting from unseen initial conditions to assess the models' generalization and long-term stability.} As shown in Fig.~\ref{fig:L63-pdf-sup}, each row corresponds to a representative modeling approach (\methodName{}, strong NODE, weak NODE, and RAFDA), and the columns show the estimated PDFs for the $x$, $y$, and $z$ state variables under 1\%, 5\%, 10\%, and 20\% noise levels. For brevity, MP-NODE and DeepSkip are omitted from this visualization, as they exhibit severe distribution collapse similar to or worse than the strong NODE under high noise. Overall, as the noise level increases, the mismatch between the predicted and true distributions becomes more significant for most methods. Notably, under 10\% and 20\% noise, both the weak and strong NODE struggle to recover meaningful distributions, often collapsing toward narrow or overly concentrated profiles that deviate substantially from the true attractor. While RAFDA demonstrates competitive visual robustness and avoids complete collapse, it exhibits noticeable offsets and wider deviations from the ground-truth peaks. In contrast, the proposed \methodName{} continues to produce PDFs that closely resemble the ground truth across all state dimensions, demonstrating strong robustness in preserving long-term statistical structure even in highly noisy conditions.

These results reinforce the KL divergence findings and further highlight the advantage of the proposed method in capturing the correct invariant measure under observational noise.

\section{KS Analysis}\label{app:ks}
\begin{figure}[!htbp]
    \centering
    \includegraphics[width=\textwidth]{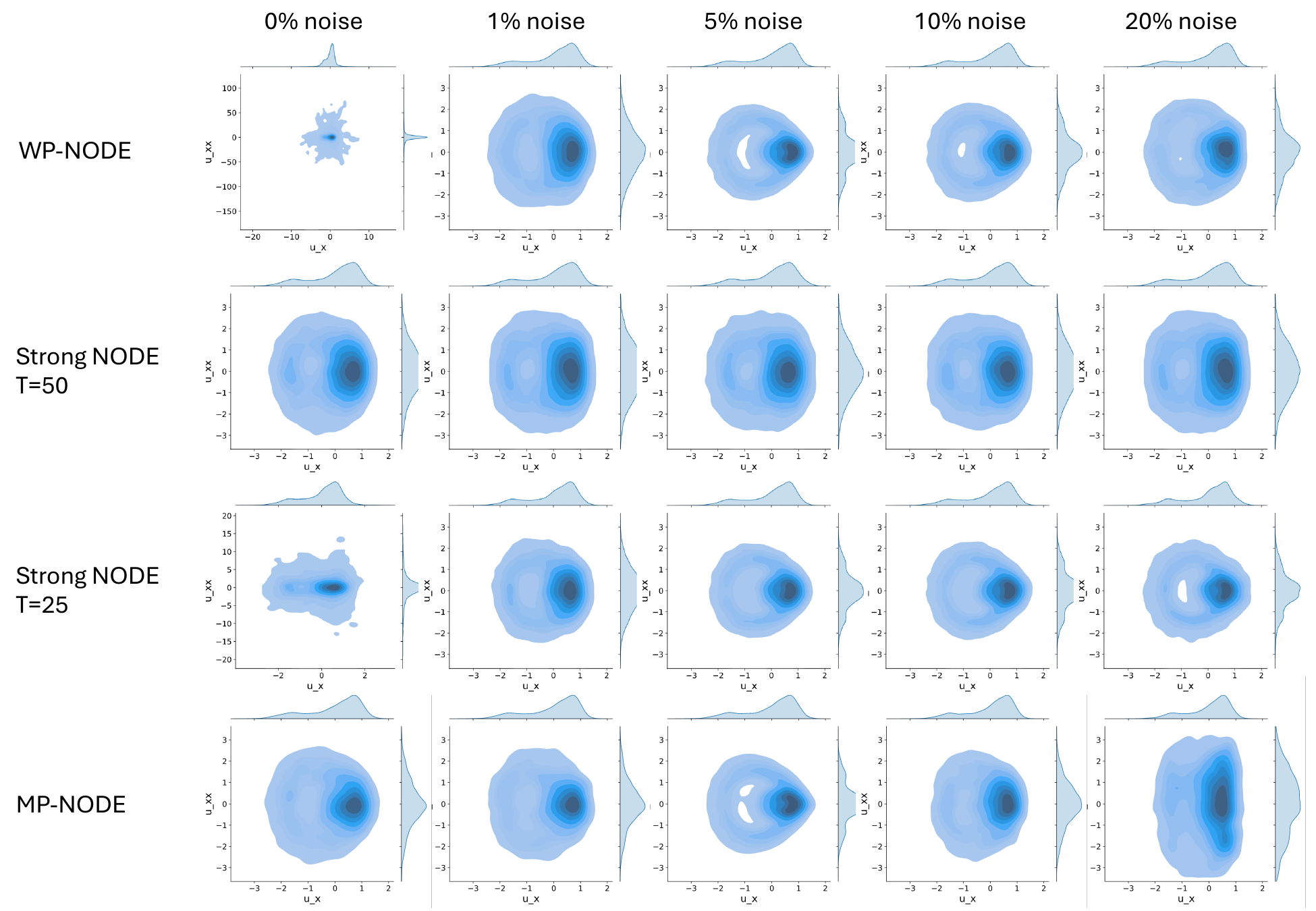}
    \caption{Invariant measure comparisons of the learned KS system across different methods under varying noise conditions.}
    \label{fig:KS-pdf-sup}
\end{figure}

This section provides additional visualizations of the long-term statistical properties of the learned KS system. Specifically, Figure~\ref{fig:KS-pdf-sup} displays the joint probability density plots for \methodName{}, MP-NODE, and two variants of the strong NODE baseline (trained with rollout lengths $T=25$ and $T=50$) across different observation noise levels (0\%, 1\%, 5\%, 10\%, and 20\%). These visualizations directly complement the quantitative KL divergence results reported in Table~\ref{tab:ks_vpt_kl} of the main text.

As observed in the figure, \methodName{} accurately reproduces the invariant measure at 5\% and 10\% noise, but degrades in the noiseless (0\%) regime.
In contrast, the strong form baseline exhibits high sensitivity to the training rollout length. While the strong NODE with $T=25$ reasonably approximates the core of the distribution, increasing the rollout to $T=50$ causes the model to almost completely fail in capturing the joint invariants, resulting in highly distorted density estimates.
Meanwhile, MP-NODE aligns well with the reference only at 5\% noise, degrading at all other levels. 
Finally, the weak NODE is omitted from these visualizations as it completely fails to produce stable long-term trajectories.

\section{ERA5 Analysis}\label{app:era5}
\begin{figure}[!htbp]
    \centering
    \includegraphics[width=0.95\textwidth]{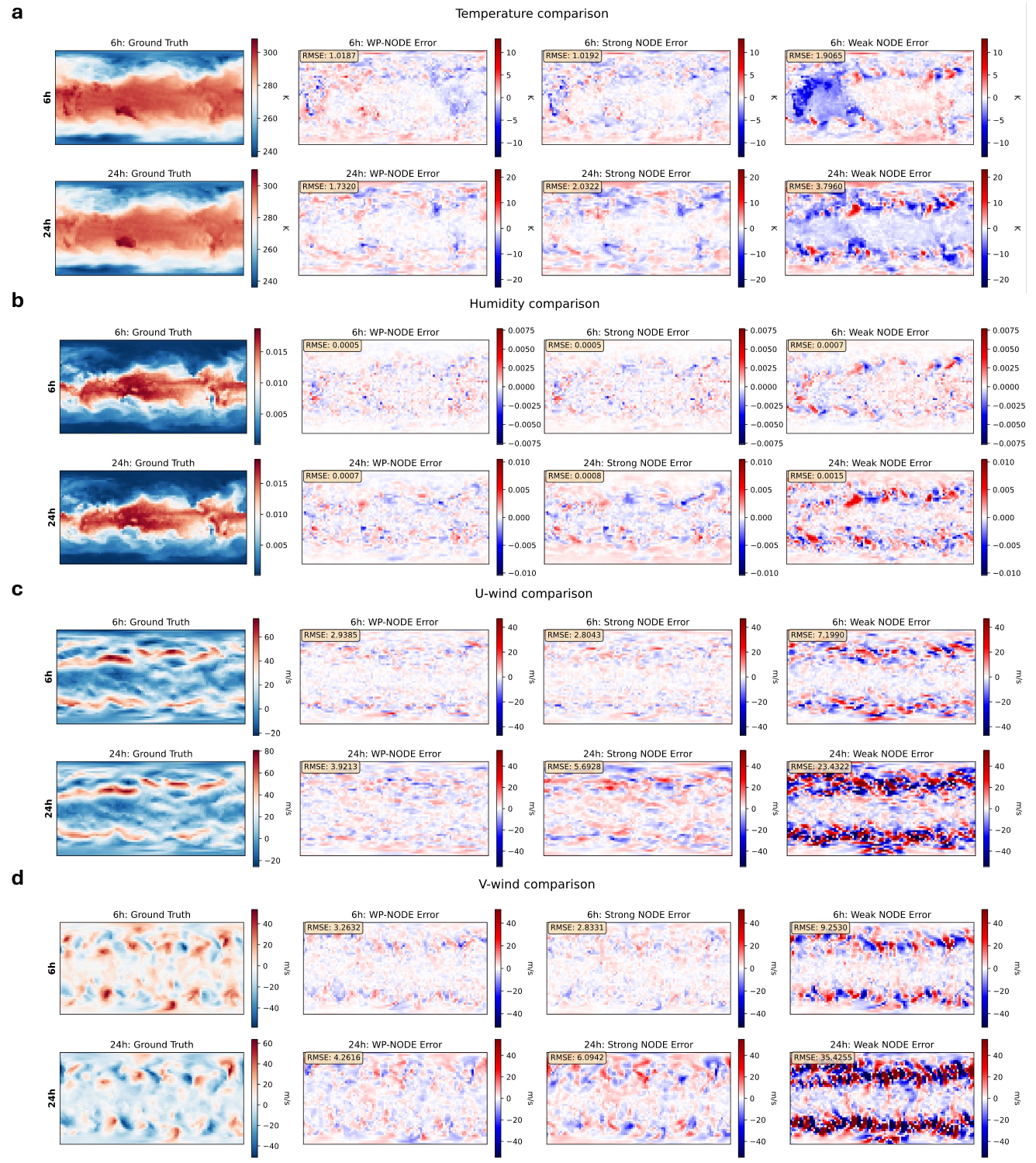}
    \caption{Short-term prediction comparison (using RMSE) on the ERA5 dataset over the four prognostic variables. \textbf{a}. Temperature. \textbf{b}. Humidity. \textbf{c}. Zonal component of the horizontal wind velocity. \textbf{d}. Meridional component of the horizontal wind velocity.}
    \label{fig:ERA5-sup}
\end{figure}

\begin{figure}[!htbp]
    \centering
    \includegraphics[width=\textwidth]{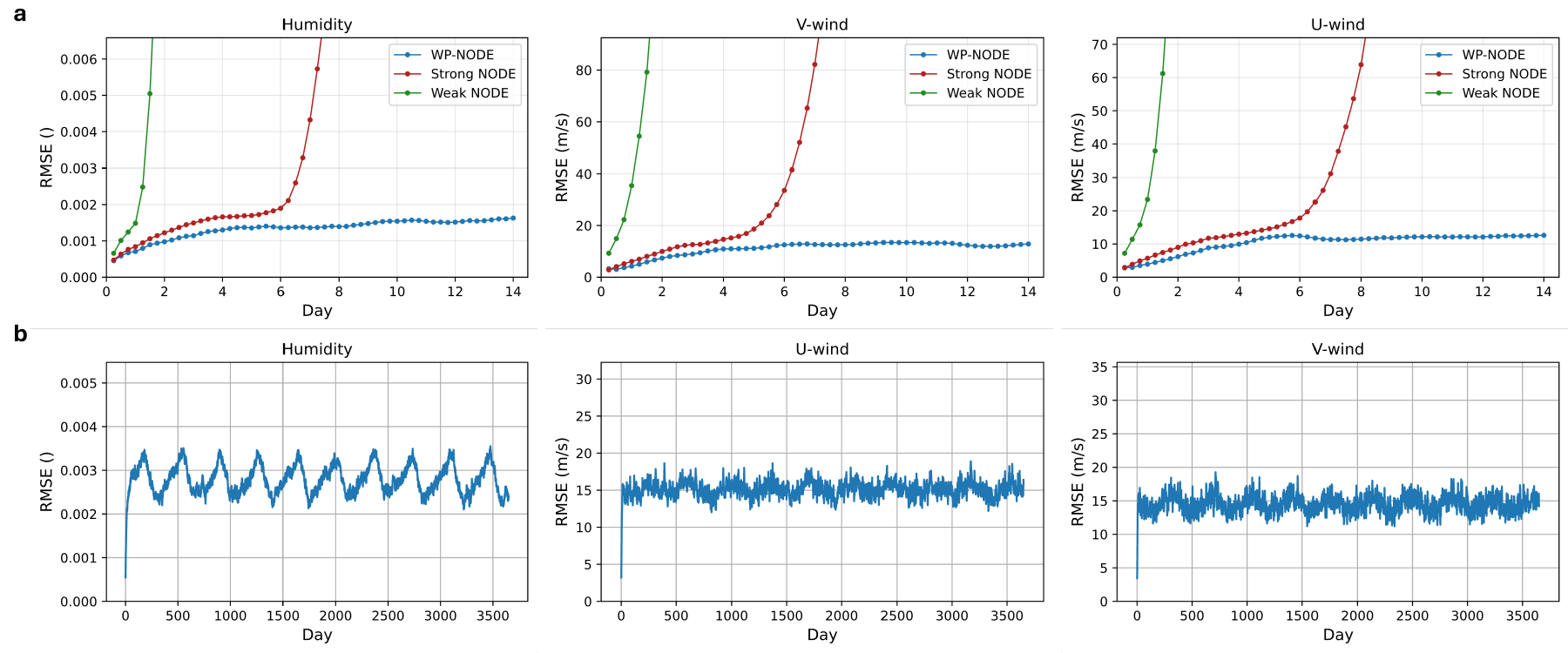}
    \caption{Extended prediction evaluation on the ERA5 dataset for specific humidity, and the zonal and meridional components of the horizontal wind velocity. \textbf{a}. 14-day RMSE evolution. \textbf{b}. 1-year prediction performance of \methodName{}.}
    \label{fig:ERA5-sup2}
\end{figure}

In Fig.~\ref{fig:ERA5-sup}, short-term predictions from \methodName{} are compared with two baseline NODE models for 6-hour and 24-hour forecasts across the four prognostic variables. \methodName{} consistently outperforms both baselines across all variables and time horizons.

In Fig.~\ref{fig:ERA5-sup2}a, the 14-day RMSE evolution is shown for the three prognostic variables not included in the main article (the temperature variable was presented in the main text). The weak and strong baseline NODE models diverge and produce unstable predictions within days, while \methodName{} maintains stable and accurate forecasts throughout the 14-day period. Fig.~\ref{fig:ERA5-sup2}b further demonstrates that \methodName{} can maintain forecast stability over significantly longer time horizons.
These results further strengthen the claims in the main article that \methodName{}'s hybrid training strategy delivers superior stability, accuracy, and long-term predictive fidelity in real-world climate forecasting, precisely where conventional NODE approaches fail. 

\section{Implementation Details of the Weak Form}\label{app:integration}
In this appendix, we provide the details for the weak formulation. Specifically, we state the polynomial test function and discuss the quadrature rule that forms the weak formulation loss function. 

\subsection{Polynomial Test Functions}
This section provides the explicit derivative expansions, antiderivatives, and weight constructions for the polynomial test functions $\phi_p(s)=(1-s^2)^p$ used in the weak form integration. These functions vanish at the segment boundaries, $\phi_p(\pm 1)=0$, which makes them particularly well-suited for localized formulations.  

The $d^{\text{th}}$ derivative of $\phi_p$ admits the closed-form expansion
\begin{align}
\phi_p^{(d)}(s)=
\sum_{k=0}^{p}
(-1)^{k}\binom{p}{k}
\frac{(2p - 2k)!}{(2p - 2k - d)!}
s^{2(p-k)-d}.\notag
\end{align}

Since $\phi_p^{(d)}(s)$ and $s\phi_p^{(d)}(s)$ are polynomials, their primitives also have closed-form expressions:
\begin{align}
\begin{aligned}
\Phi_{p,d}(s) &= \int \phi_p^{(d)}(s)ds, \\
\Psi_{p,d}(s) &= \int s\phi_p^{(d)}(s)ds.
\end{aligned}\notag
\end{align}

The finite weights used in the weak form summation are constructed from differences of these antiderivatives evaluated at segment boundaries. 

\subsection{Piecewise Linear Interpolation and Decoupling of Integral} 
This subsection provides the detailed derivation of the weight vectors $w_{\text{lhs},i}$ and $w_{\text{rhs},i}$ used in the discrete weak form. The interpolation basis is first defined, followed by a demonstration of how the weak form integrals decompose into weighted sums of nodal values.

To evaluate the weak form in practice, the reference domain $s \in [-1,1]$ of each segment is utilized and discretized with $\ell+1$ grid points $s_i$. Let \( s_0 < s_1 < \dots < s_{\ell} \) be partition points on the interval \([s_0, s_{\ell}]\), and suppose values \( u_i = u(s_i) \) and \( f_i = f(u_i,s_i;\theta)\) for \( i = 0, \dots, \ell \) are given.

\noindent \textbf{1. Linear Basis Functions.}
Define the piecewise linear Lagrange basis functions \( L_i(s) \),
\begin{equation}
L_i(s) = 
\begin{cases}
\displaystyle \frac{s - s_{i-1}}{s_i - s_{i-1}}, & s \in [s_{i-1}, s_i] \\[3ex]
\displaystyle \frac{s_{i+1} - s}{s_{i+1} - s_i}, & s \in [s_i, s_{i+1}] \\[3ex]
0, & \text{otherwise}
\end{cases}
\quad \text{for } 1 \le i \le \ell-1. \notag
\end{equation}

For the endpoints:
\begin{align}
L_0(s) = 
\begin{cases}
\displaystyle \frac{s_1 - s}{s_1 - s_0}, & s \in [s_0, s_1] \\[3ex]
0, & \text{otherwise}
\end{cases}
\quad\quad
L_{\ell}(s) = 
\begin{cases}
\displaystyle \frac{s - s_{\ell-1}}{s_{\ell} - s_{\ell-1}}, & s \in [s_{\ell-1}, s_{\ell}] \\[3ex]
0, & \text{otherwise}
\end{cases}\notag
\end{align}

\noindent \textbf{2. Interpolant Construction.}
The linear interpolants of the functions \( u(s) \) and $f(u(s),s; \theta)$ are defined as:
\begin{align}
\tilde{u}(s) = \sum_{i=0}^{\ell} u_i  L_i(s), \quad\quad 
\tilde{f}(s) = \sum_{i=0}^{\ell} f_i  L_i(s). \label{eq:lin} 
\end{align}

These functions are continuous and piecewise linear. Note that if $u_i, f_i \in \mathbb{R}^D$ are vectors and $L_i(s) \in \mathbb{R}$ is a scalar, the terms $u_iL_i(s)$ and $f_iL_i(s)$ represent the standard scalar-vector products.

\noindent \textbf{3. Projection Against a Test Function.}
The following integrals are approximated using Eq.~\eqref{eq:lin} as follows,
\begin{align}
\frac{\ell\Delta t}{2}\int_{s_0}^{s_{\ell}} f(u(s),s;\theta) \phi(s)  ds \approx \frac{\ell\Delta t}{2}\int_{s_0}^{s_{\ell}} \tilde{f}(s) \phi(s)  ds = \sum_{i=0}^{\ell} f_i  \left( \frac{\ell\Delta t}{2}\int_{s_0}^{s_{\ell}} L_i(s) \phi(s)  ds \right)
= \sum_{i=0}^{\ell} f_i w_{\text{rhs},i},\notag
\end{align}
where the weights are defined as:
\begin{align}
w_{\text{rhs},i} = \frac{\ell\Delta t}{2}\int_{s_0}^{s_{\ell}} L_i(s)  \phi(s)  ds.\notag
\end{align}

Similarly,
\begin{align}
\int_{s_0}^{s_{\ell}} u(s) \phi'(s)  ds 
\approx \int_{s_0}^{s_{\ell}} \tilde{u}(s) \phi'(s)  ds =  \sum_{i=0}^{\ell} u_i \left(\int_{s_0}^{s_{\ell}} L_i(s) \phi'(s)  ds\right)=  \sum_{i=0}^{\ell} u_i w_{\text{lhs},i},\notag
\end{align}

where
\begin{align}
w_{\text{lhs},i} := \int_{s_0}^{s_{\ell}} L_i(s) \phi'(s)  ds.\notag
\end{align}

These weights can be precomputed for a fixed \( \phi(s) \), independent of \( u \).

\noindent \textbf{4. Explicit Expression for Weights.}
For interior indices \( 1 \le i \le \ell-1 \), the weights are given by:
\begin{align}
\begin{aligned}
w_{\text{rhs},i} &= \frac{\ell\Delta t}{2}
\int_{s_{i-1}}^{s_i} \frac{s - s_{i-1}}{s_i - s_{i-1}}  \phi(s)  ds 
+ \frac{\ell\Delta t}{2} \int_{s_i}^{s_{i+1}} \frac{s_{i+1} - s}{s_{i+1} - s_i}  \phi(s)  ds \\
&= \frac{\ell\Delta t}{2(s_i - s_{i-1})} \left[ \Psi(s_i) - \Psi(s_{i-1}) - s_{i-1}  (\Phi(s_i) - \Phi(s_{i-1})) \right] \notag \\ &\quad + \frac{\ell\Delta t}{2(s_{i+1} - s_i)} \left[ s_{i+1}  (\Phi(s_{i+1}) - \Phi(s_i)) - \Psi(s_{i+1}) + \Psi(s_i) \right],
\end{aligned}\notag
\end{align}

where $\Phi_{p,d}(s)=\!\int\!\phi_p^{(d)}(s)ds$ and $\Psi_{p,d}(s)=\!\int\!s\phi_p^{(d)}(s)ds$.

For the endpoints:
\begin{align}
\begin{aligned}
w_{\text{rhs},0} &= \frac{\ell\Delta t}{2}\int_{s_0}^{s_1} \frac{s_1 - s}{s_1 - s_0} \phi(s)  ds = \frac{\ell\Delta t}{2(s_1 - s_0)} \left[ s_1  (\Phi(s_1) - \Phi(s_0)) - \Psi(s_1) + \Psi(s_0) \right],
\end{aligned}\notag
\end{align}

and,
\begin{align}
\begin{aligned}
w_{\text{rhs},\ell} &= \frac{\ell\Delta t}{2}\int_{s_{\ell-1}}^{s_{\ell}} \frac{s - s_{\ell-1}}{s_{\ell} - s_{\ell-1}} \phi(s)  ds \\
&= \frac{\ell\Delta t}{2(s_{\ell} - s_{\ell-1})} \left[ \Psi(s_{\ell}) - \Psi(s_{\ell-1}) - s_{\ell-1}  (\Phi(s_{\ell}) - \Phi(s_{\ell-1})) \right].
\end{aligned}\notag
\end{align}

The other weights can also be computed as follows:
\begin{align}
\begin{aligned}
w_{\text{lhs},i} &=
\int_{s_{i-1}}^{s_i} \frac{s-s_{i-1}}{s_i-s_{i-1}}\phi'(s)ds +
\int_{s_i}^{s_{i+1}} \frac{s_{i+1}-s}{s_{i+1}-s_i}\phi'(s)ds \\[6pt]
&=
\Bigl[\tfrac{s-s_{i-1}}{s_i-s_{i-1}}\phi(s)\Bigr]_{s_{i-1}}^{s_i}
-\frac{1}{s_i-s_{i-1}}\int_{s_{i-1}}^{s_i}\phi(s)ds +
\Bigl[\tfrac{s_{i+1}-s}{s_{i+1}-s_i}\phi(s)\Bigr]_{s_i}^{s_{i+1}}
+\frac{1}{s_{i+1}-s_i}\int_{s_i}^{s_{i+1}}\phi(s)ds \\[8pt]
&=
-\frac{s_{i-1}}{s_i-s_{i-1}}\bigl[\Phi(s_i)-\Phi(s_{i-1})\bigr]
+\frac{1}{s_i-s_{i-1}}\bigl[\Psi(s_i)-\Psi(s_{i-1})\bigr] \\[4pt]
&\quad+
\frac{s_{i+1}}{s_{i+1}-s_i}\bigl[\Phi(s_{i+1})-\Phi(s_i)\bigr]
-\frac{1}{s_{i+1}-s_i}\bigl[\Psi(s_{i+1})-\Psi(s_i)\bigr].
\end{aligned}\notag
\end{align}

For the endpoints, it similarly follows that:
\begin{align}
\begin{aligned}
w_{\text{lhs},0} &= 
\frac{s_1}{s_1 - s_0}\bigl[\Phi(s_1) - \Phi(s_0)\bigr] \;-\;
\frac{1}{s_1 - s_0}\bigl[\Psi(s_1) - \Psi(s_0)\bigr], \\[6pt]
w_{\text{lhs},\ell} &= 
-\frac{s_{\ell-1}}{s_{\ell} - s_{\ell-1}}\bigl[\Phi(s_{\ell}) - \Phi(s_{\ell-1})\bigr] \;+\;
\frac{1}{s_{\ell} - s_{\ell-1}}\bigl[\Psi(s_{\ell}) - \Psi(s_{\ell-1})\bigr].
\end{aligned}\notag
\end{align}

\subsection{Vectorized Weak Form Loss}
To effectively evaluate the weak form in practice, the computation is highly vectorized, partially following the weak-SINDy implementation. Both the data $u(s)$ and the function evaluations $f\big(u(s),s;\theta\big)$ are represented using piecewise linear interpolation between $\ell+1$ grid points on each segment. This allows the weak form integrals to be expressed as linear combinations of the nodal values $u_i$ and the neural network evaluations at those nodes $f_i:=f(u_i, t_i;\theta)$. Particularly,

\begin{align}
\begin{aligned}
\int_{-1}^{1} u(\tau_{k,\ell}(s))\phi_p'(s)ds 
&\approx \sum_{i=0}^{\ell} u_i w_{\text{lhs},i} 
= \mathbf{u}^\top \mathbf{w}_{\text{lhs}}, \\[6pt]
\frac{\ell\Delta t}{2}\int_{-1}^{1} f(u(\tau_{k,\ell}(s)),\tau_{k,\ell}(s);\theta)\phi_p(s)ds 
&\approx \sum_{i=0}^{\ell}f_i w_{\text{rhs},i} 
= \mathbf{f}^\top \mathbf{w}_{\text{rhs}},\notag
\end{aligned}
\end{align}
where the precomputable weights are $w_{\text{lhs},i}=  \int L_i(s)\phi'(s)ds$ 
and $w_{\text{rhs},i}=\frac{\ell\Delta t}{2}\int L_i(s)\phi(s)ds$. Here,
$L_i(s)$ denotes the linear Lagrange basis function centered at the grid point $s_i$.

Since $\phi_p$ is smooth and $L_i(s)$ is linear, the weights admit exact closed-form expressions. In vectorized form, by substituting the noisy observations $v$ for the unobserved true states $u$ (as noted in Section~\ref{sec22}), the weak form evaluation for a single segment $k$ is defined as the squared residual:
\begin{align}\label{eq:weak_loss}
\mathcal{L}_{\text{weak}}(\theta)=\| V_{k,\ell}^p + F^p_{k,\ell}(\theta) \|^2
= \left\|\mathbf{v}^\top \mathbf{w}_{\text{lhs}}
+ \mathbf{f}^\top \mathbf{w}_{\text{rhs}}\right\|^{2},
\end{align}
where $\mathbf{f}$ denotes the vector of neural network evaluations at the nodal points $u$ in each segment, i.e., $\mathbf{f} = \bigl[f(v_0, t_0;\theta), f(v_1, t_1;\theta), \dots, f(v_{\ell}, t_{\ell};\theta)\bigr]^\top$. This formulation naturally generalizes to higher-dimensional or vector-valued data via batched dot products across dimensions. {\color{black}Crucially, since Eq.~\eqref{eq:weak_loss} bypasses ODE integration, $\mathcal{L}_{\text{weak}}$ is optimized via standard backpropagation rather than backpropagation through ODE solvers, significantly reducing memory and time overhead.}

\section{Comparison with VF-NODE}\label{app:vfnode}

\begin{sloppypar}
VF-NODE~\cite{zhao2025accelerating} trains NODE with a variational loss built from a global Fourier sine basis, $\phi_m(t) = \sqrt{2/T_{\mathrm{total}}}\sin(\pi m t/T_{\mathrm{total}}), m = 1,\ldots,L_b$, where $L_b$ is the number of basis functions and $T_{\mathrm{total}}$ denotes the total length of the training time series. The noisy trajectory is first represented by a natural cubic spline, and the resulting closed-form approximation is integrated against each $\phi_m$ via Filon quadrature. 
\end{sloppypar}

All ODE-simulation benchmarks in \cite{zhao2025accelerating} are non-chaotic systems observed on short windows ($T_{\mathrm{total}}=10$, roughly $100$ points per trajectory). Their Appendix~J reports a Lorenz experiment at $\rho=20$, below the standard chaotic regime, evaluated via unstable-periodic-orbit detection rather than trajectory metrics. 

We implemented VF-NODE on the Lorenz-63 setup of Section~\ref{sec41} ($\rho = 28$, $T_{\mathrm{total}}=100$\,s, $N=10^4$), using the same network, data, and evaluation protocol as all baselines in Table~\ref{tab:cross_system_vpt_kl}, with a training budget of up to $20{,}000$ epochs with early stopping (four times the default of \cite{zhao2025accelerating}). The implementation was validated against analytic test functions with known exact derivatives, for which the variational residual is at the $10^{-6}$ level. The spline and Filon kernels are evaluated in double precision, since in single precision the low-frequency integrals suffer catastrophic cancellation that corrupts precisely the constraints encoding the slow dynamics; \methodName{} and all other baselines use single precision throughout, so this treatment is applied to VF-NODE alone, in its favor.

{\color{black}
With the settings reported in~\cite{zhao2025accelerating} ($\lambda=0.99$, $L_b=80$), VF-NODE fails entirely on this chaotic system, reaching VPT $<0.2$ Lyapunov times and a diverging long-term rollout. We therefore swept $\lambda$ and $L_b$ jointly; Tables~\ref{tab:vfnode_lambda} and~\ref{tab:vfnode_Lb} report the two one-dimensional slices through the best-performing configuration, $\lambda = 0.9999$ and $L_b = 640$, whereas the ablation study of \cite{zhao2025accelerating} considers at most $L_b=110$.

\begin{table}[!htbp]
\centering
\renewcommand{\ms}[2]{#1{\scriptsize$\pm$}\scalebox{0.9}{#2}}
\caption{Effect of the spline smoothing parameter $\lambda$ on VF-NODE for L63 ($5\%$ noise, $L_b=640$); $\lambda=0.99$ is the default of~\cite{zhao2025accelerating}.}
\label{tab:vfnode_lambda}
\begin{tabular}{lcc}
\toprule
$\lambda$ & VPT $\uparrow$ & KL $\downarrow$ \\
\midrule
$0.99$    & \ms{0.161}{0.119} & -- \\
$0.995$   & \ms{0.234}{0.165} & -- \\
$0.999$   & \ms{0.704}{0.812} & -- \\
$0.9999$  & \ms{0.860}{0.533} & -- \\
$0.99999$ & \ms{0.993}{0.594} & -- \\
\bottomrule
\end{tabular}
\end{table}

\begin{table}[!htbp]
\centering
\renewcommand{\ms}[2]{#1{\scriptsize$\pm$}\scalebox{0.9}{#2}}
\caption{Effect of the basis size $L_b$ on VF-NODE for L63 ($5\%$ noise, $\lambda=0.9999$). $L_b=80$ is the default of~\cite{zhao2025accelerating}.}
\label{tab:vfnode_Lb}
\begin{tabular}{lcc}
\toprule
$L_b$ & VPT $\uparrow$ & KL $\downarrow$ \\
\midrule
$40$   & \ms{0.094}{0.061} & -- \\
$80$   & \ms{0.100}{0.056} & -- \\
$160$  & \ms{0.145}{0.078} & 13.259 \\
$320$  & \ms{0.322}{0.237} & 6.437 \\
$640$  & \ms{0.860}{0.533} & -- \\
\bottomrule
\end{tabular}
\end{table}

Table~\ref{tab:vfnode_lambda} shows that the smoothing parameter $\lambda$ dominates the outcome: at the default $\lambda=0.99$ the model fails to recover the dynamics, and performance improves with $\lambda$ but with only marginal gains beyond $\lambda = 0.9999$. We use $\lambda = 0.9999$ in what follows.
At this $\lambda$, Table~\ref{tab:vfnode_Lb} shows that VF-NODE improves with $L_b$, but even at $L_b=640$ its VPT remains well below the other NODE variants. This reflects a structural constraint: the finest test function $\sin(\pi L_b t/T_{\mathrm{total}})$ resolves frequencies up to $L_b/(2T_{\mathrm{total}})$, so $L_b$ must grow linearly with the length of the training time series. On the KS time series of Section~\ref{sec52} ($T_{\mathrm{total}} = 25{,}000$\,s) the required basis size therefore becomes prohibitive, and the loss remains a single global integral that cannot be mini-batched, unlike the locally supported test functions of \methodName{}. 
Since learning the invariant measures of chaotic systems inherently requires long time series, this scaling negates the computational advantage of VF-NODE precisely in the regime addressed in this paper. In contrast, the resolution of the locally supported test functions of \methodName{} is independent of the length of the training time series. In both tables, the KL divergence does not follow the VPT trend, since configurations with better VPT do not necessarily produce bounded long-term rollouts.
}


\end{document}